%% file: Arxiv.tex
\documentclass[journal]{IEEEtai}

\usepackage[colorlinks,urlcolor=blue,linkcolor=blue,citecolor=blue]{hyperref}

\usepackage{color,array}

\usepackage{graphicx}

\usepackage{amsmath,amsfonts}
\usepackage{algorithmic}
\usepackage{algorithm}
\usepackage{array}
\usepackage[caption=false,font=normalsize,labelfont=sf,textfont=sf]{subfig}
\usepackage{textcomp}
\usepackage{stfloats}
\usepackage{url}
\usepackage{verbatim}
\usepackage{graphicx}
\usepackage{cite}
\usepackage[colorinlistoftodos]{todonotes}

\usepackage{tikz}
\usepackage{forest}

\usetikzlibrary{trees,positioning,shapes,shadows,arrows.meta, backgrounds}

\usepackage[normalem]{ulem}  
\usepackage{pgfplots}
\pgfplotsset{compat=1.18}
\usepackage{amsmath}

\usepackage{tabularx}
\usepackage{booktabs}
\usepackage{makecell}
\newcolumntype{Y}{>{\raggedright\arraybackslash}X}

\definecolor{frozenBlue}{RGB}{200, 220, 245}
\definecolor{frozenBorder}{RGB}{60, 90, 140}
\definecolor{trainOrange}{RGB}{255, 230, 200}
\definecolor{trainBorder}{RGB}{200, 100, 0}
\definecolor{gradRed}{RGB}{200, 40, 40}
\definecolor{flowGreen}{RGB}{30, 130, 30}
\definecolor{roundingBorder}{RGB}{165, 177, 194}
\definecolor{roundingGrey}{RGB}{209, 216, 224}

\usepackage[dvipsnames]{xcolor}

\usepackage{booktabs, threeparttable, array} 

\begin{document}

\title{Unifying Data, Memory, and Compute Efficiency in LLM training: A Survey} 

\author{Vanessa Schmidt, Huy Hoang Nguyen, Cédric Jung,  Shirin Salehi, and Anke Schmeink
\thanks{
}

\thanks{V. Schmidt, S. Salehi, and A. Schmeink are with the
Chair of Information Theory and Data Analytics (INDA), RWTH Aachen University,
52074 Aachen, Germany. (e-mail: vanessa.schmidt@rwth-aachen.de; shirin.salehi@inda.rwth-aachen.de; anke.schmeink@inda.rwth-aachen.de). 

Huy Hoang Nguyen and C. Jung are with the AIT Austrian Institute of Technology GmbH, 1210 Vienna, Austria. C. Jung is also with the Automation and Control Institute, Technische Universit\"at Wien (TUW), 1040 Vienna, Austria. (e-mail: huy-hoang.nguyen@ait.ac.at; jung@acin.tuwien.ac.at)}
}


\maketitle

\begin{abstract}
Resource constraints increasingly determine what can be trained, fine-tuned, and deployed in large language models (LLMs), yet efficiency is often studied through isolated techniques rather than as an interacting system of limits. This survey adopts a constraint-centric perspective and organizes recent progress around three coupled bottlenecks: data efficiency (what to train on), memory efficiency (how to fit training), and compute budget awareness (when and where to spend FLOPs).
On the data axis, we review selection and pruning methods that maximize learning per token, ranging from scalable proxy signals based on learning dynamics to gradient- and influence-based scoring, as well as difficulty-aware and curriculum-style strategies. We highlight emerging evidence that different notions of “good data” dominate in different regimes, implying that optimal subsets depend on the task objective and resource budget rather than being universal.
On the systems side, we show that GPU memory, not raw compute, is often the dominant bottleneck in fine-tuning, and that effective scaling requires jointly reducing weight storage, optimizer states, and activation memory rather than optimizing any single component in isolation. Beyond memory, we frame training and inference as compute-governed processes in which optimization, data selection, and decoding must explicitly account for finite FLOP budgets. We review evidence for compute-optimal allocation and stopping rules, where computation should be halted or reallocated once marginal performance gains fall below a budget-dependent threshold. Together, these results unify compute-aware data selection, scaling laws, and adaptive inference under a common principle of resource-conditioned decision-making.

\end{abstract}

\begin{IEEEImpStatement}
Large language models (LLMs) offer transformative capabilities but remain challenging to train and deploy on resource-constrained devices due to their data, memory, and compute demands. This paper surveys existing techniques and organizes them along a resource-constrained lifecycle, addressing what to train (data), how to fit it (memory), and when to stop or reallocate (compute). By highlighting the trade-offs and interdependencies across these dimensions, we reveal how isolated optimizations often shift bottlenecks rather than resolve them. We also identify a critical gap: the predominance of static, pre-filtered data selection methods and the need for dynamic, influence-aware approaches during training. These insights provide a foundation for energy-efficient, edge-compatible LLMs, reducing operational cost and environmental impact, and enabling LLMs to run sustainably on mobile, industrial, and wearable platforms.
\end{IEEEImpStatement}


\section{Introduction}

\IEEEPARstart{I}{n} recent years, large language models (LLMs) have redefined the frontier of artificial intelligence (AI) by achieving exceptional performance in natural language understanding, generation, and complex reasoning across diverse application domains. This progress has been largely driven by the rapid scaling of model size and context length in prominent models such as GPT-3 with 175 billion parameters, Google’s PaLM, a large transformer trained using the Pathways system with 540 billion parameters~\cite{Bai2024Survey, Chowdhery2022PaLM}. However, this rapid advancement has been accompanied by a drastic increase in computational, memory, energy, and financial costs, driven not only by ever-larger parameter counts and extended context windows that are central to modern deep learning architectures, but also by the need for abundant, high-quality training data~\cite{Samsi2023Energy}. The resource footprint of training and deploying such models, including GPU hours, energy consumption, carbon emissions, and water usage, poses significant barriers for researchers and practitioners, particularly in resource-constrained environments such as academic laboratories, edge devices, or critical sectors like healthcare and finance~\cite{Jegham2025Hungry, GreenAI2024}, as well as environmental sustainability\cite{salehi2023data}. Tackling these challenges requires comprehensive strategies to enhance the resource efficiency of LLMs from training to deployment~\cite{Bai2024Survey}.
This is particularly vital for edge
intelligence, where the ``optimization trilemma" of data, memory, and compute is not just a theoretical framework but a physical necessity. For instance, on mobile or industrial edge devices, aggressive data pruning (data efficiency) can reduce training iterations (compute efficiency), which directly translates to lower thermal output and battery preservation, factors as critical as the model's accuracy itself.

\begin{figure*}[t!]
\centering
\resizebox{0.95\textwidth}{!}{%
    \begin{tikzpicture}[
        font=\sffamily,
        >=Stealth,
        node distance=1.0cm and 2.5cm, 
        process/.style={
            rectangle, 
            draw=frozenBorder, 
            thick, 
            fill=frozenBlue!20, 
            rounded corners, 
            minimum width=4cm, 
            minimum height=2.2cm, 
            align=center,
            drop shadow
        },
        labelnode/.style={
            font=\bfseries\small,
            text=black!80
        },
        arrow/.style={
            ->, 
            thick, 
            color=black!70,
            text=black,
            font=\footnotesize
        },
        arrowtext/.style={
            midway,
            above,
            align=center,
            font=\normalsize, 
            text width=2cm    
        }
    ]


    \node[process] (data) {
        \textbf{III. Data Efficiency}\\
        \textit{(The Foundation)}\\
        \rule{3.5cm}{0.5pt}\\
        \small{Selection \& Pruning}
    };

    \node[process, right=of data] (memory) {
        \textbf{IV. Memory Efficiency}\\
        \textit{(The Constraint)}\\
        \rule{3.5cm}{0.5pt}\\
        \small{Fitting the Training}
    };

    \node[process, right=of memory] (compute) {
        \textbf{V. Compute Budget}\\
        \textit{(The Governor)}\\
        \rule{3.5cm}{0.5pt}\\
        \small{Budget Allocation}
    };

    \node[left=0.8cm of data, font=\bfseries] (input) {Raw Data};
    \node[right=0.8cm of compute, font=\bfseries, align=center] (output) {LLM};


    \draw[arrow] (input) -- (data);
    
    \draw[arrow] (data) -- node[arrowtext] {High-Utility\\Data} (memory);
    \draw[arrow] (memory) -- node[arrowtext] {Feasible\\Step} (compute);
    
    \draw[arrow] (compute) -- (output);


    \draw[arrow, dashed, color=gradRed] (compute.south) -- ++(0,-1.0) -| node[pos=0.25, above, color=gradRed, font=\normalsize] {Stop or Re-evaluate} (data.south);

    \end{tikzpicture}
}
\caption{\textbf{The Resource-Constrained Lifecycle and Survey Structure.} This figure illustrates the unified framework proposed in this survey. Section \ref{data} (Data Efficiency) determines the high-utility inputs, Section \ref{memory} (Memory Efficiency) addresses the feasibility of fitting training into hardware constraints, and Section \ref{compute} (Compute Budget) acts as the governor for resource allocation. Dashed lines represent the coupled feedback loops essential for true efficiency.}
\label{fig:unified_lifecycle_simple}
\end{figure*}
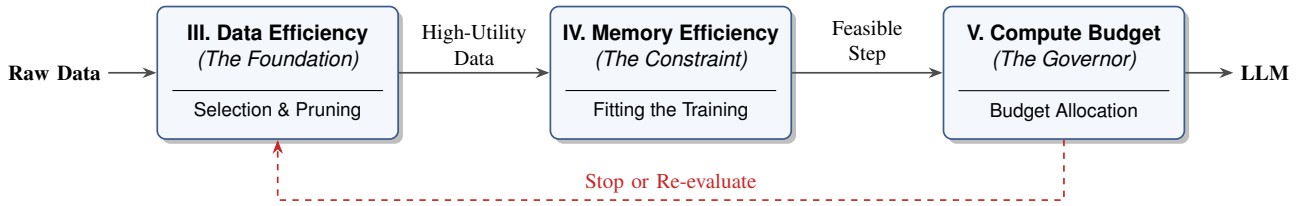


Prior work on LLM efficiency has extensively studied data-centric approaches, motivated by empirical and theoretical scaling laws showing diminishing returns from indiscriminate data scaling~\cite{sorscher2022beyond}. Numerous studies demonstrate that training corpora contain substantial redundancy~\cite{yang2024smalltolarge, wang2024greats}, task misalignment ~\cite{du2025disentangling, xia2024less}, and imbalanced difficulty~\cite{dai2025bids}. As a result, methods such as data filtering, curriculum learning, and influence-based data selection~\cite{yincompute, yu2025llm, zhang2025staff, wang2025dynamic} aim to improve convergence and downstream performance by prioritizing informative samples.

Numerous studies have examined the interaction between the training and adaptation of LLMs and hardware memory constraints. Several studies show that large-scale optimization is often limited not by floating-point operations (FLOPs), but by the memory required to store activations and optimizer states~\cite{pudipeddi2020training}. In particular, conventional batching strategies become increasingly memory-inefficient for long-context sequences \cite{nguyen2025minibatch, li2024addax}, while commonly used optimizers introduce significant state overhead that hinders scaling to billions of parameters on commodity hardware \cite{liu2024hift, luo2024badam, zhao2024galore}. In addition, the reliance on high-precision backpropagation necessitates storing extensive activation tensors, which can render standard training pipelines impractical even when model parameters fit in memory \cite{yu2024subzero, chen2024enhancing, dao2022flashattention}. To address these challenges, existing approaches have explored alternatives to conventional optimization workflows, including block-wise parameter updates \cite{luo2024badam}, low-rank gradient projections \cite{zhao2024galore}, zeroth-order optimization methods \cite{yang2024adazeta}, and direct low-precision training schemes \cite{zhao2024direct}. 


Another line of prior work has addressed compute efficiency in large-scale model training, fine-tuning and inference, examining how computational cost scales with model size and dataset volume. Empirical scaling laws show that as models and corpora grow to trillions of parameters and tokens, training and inference are increasingly constrained by the required number of FLOPs rather than storage capacity \cite{kaplan2020scaling, hoffmann2022training, muennighoff2023scaling}. In response, several approaches explore mechanisms for reducing or reallocating computation during learning and generation. These include dynamic parameter activation through Mixture-of-Experts (MoE) architectures, which execute only a subset of model parameters per token to increase capacity without proportional compute cost \cite{jiang2024mixtralexperts,dai2024deepseekmoeultimateexpertspecialization}; token-level selection methods such as Rho-1, which prioritize high-impact tokens during training to improve learning efficiency per FLOPs~\cite{lin2025rho1tokensneed}; and adaptive inference techniques like speculative decoding, which accelerate generation by verifying inexpensive draft tokens in parallel with a larger model \cite{leviathan2023fast,chen2023accelerating}. Collectively, these methods aim to reduce training and inference cost while maintaining competitive model performance.

Taken together, existing work on data, memory, and compute efficiency has produced a rich set of techniques for reducing the resource cost of training and deploying LLMs. However, these lines of research are largely developed and evaluated in isolation, often optimizing a single resource dimension while implicitly assuming others to be unconstrained. In practice, improvements in one dimension frequently introduce new bottlenecks elsewhere, for example, sophisticated data selection methods can incur prohibitive memory or runtime overhead, while memory-saving optimization strategies may restrict data adaptivity or increase computational cost. This fragmentation obscures the underlying trade-offs faced by resource-constrained systems and limits the ability to reason about efficiency holistically. These observations motivate a unified perspective in which efficiency is treated as a coupled, budget-aware decision problem spanning data selection, memory usage, and computational allocation. We visualize this proposed framework and the organization of this survey in Figure \ref{fig:unified_lifecycle_simple}. More specifically, the contribution of this paper is listed below:

\begin{itemize}


\item Structuring the literature via the \textit{resource-constrained lifecycle}. Unlike prior surveys that organize methods by architecture or task, we structure the literature along the logical decision flow of a resource-constrained system:
\begin{itemize}
    \item The foundation (Data): Identifying what to train on by moving from simple redundancy removal to sophisticated, gradient-based influence estimation.
    \item The constraint (Memory): Solving how to fit these sophisticated selection and training methods into limited hardware via mini-batch coresets (CoLM~\cite{nguyen2025minibatch}) and optimizer fragmentation (HiFT~\cite{liu2024hift}).
    \item The governor (Compute): Deciding when to stop or switch strategies using e.g., bilevel optimization or budget-aware scaling laws.
\end{itemize}
\item Identifying the ``Static-to-Dynamic" Gap. We introduce a new perspective on the evolution of data selection. We highlight that current state-of-the-art methods (like LESS~\cite{xia2024less}) are predominantly static (pre-filtering) due to computational costs. We identify a critical research gap: the need for dynamic systems that re-evaluate data influence during training. We propose that bridging this gap requires hybridizing data selection with memory-efficient approximations (e.g., using CoLM-style coresets to approximate dynamic influence cheaply), effectively moving toward a ``Dynamic LESS."
\item Unifying principles through \textit{marginal utility}. We synthesize diverse techniques under the shared principle of marginal utility per resource. Whether it is filtering a data sample, adding a parameter, or extending training by an epoch, we show that the unifying goal across all surveyed methods is to maximize performance gain per unit of constrained resource (GB VRAM, FLOPs, or Wall-time). This distinction helps practitioners separate methods that merely make training feasible (feasibility) from those that make it compute-optimal (optimality).



\end{itemize}

The organization of this paper is as follows: Section~\ref{preliminary} discusses the background and preliminaries of LLMs, pretraining, and fine-tuning. Section~\ref{data} introduces data efficiency techniques, including data selection, pruning, and curriculum strategies. Section~\ref{memory} then covers memory efficiency methods such as parameter-efficient fine-tuning (PEFT) and quantization. Section~\ref{compute} presents compute budget awareness, including compute-optimal scaling laws, compute-aware data selection, and budget-aware inference and memory-compute trade-offs. Finally, Section~\ref{conclusion} concludes the paper. 

\section{Background and Preliminaries on Large Language Models}\label{preliminary}

\subsection{Architectures and Training Paradigms of Large Language Models}

LLMs are neural networks designed to model and generate natural language by learning statistical regularities from large-scale text corpora. Modern LLMs are predominantly based on the Transformer architecture~\cite{vaswani2017attention}, whose central component is the self-attention mechanism. Unlike recurrent or convolutional models, self-attention enables the model to process all tokens in a sequence simultaneously, capturing long-range dependencies and complex contextual relationships with high computational parallelism.

At the core of transformer-based LLMs lies the concept of token embeddings. Since neural networks cannot directly process symbolic text, discrete tokens are mapped into continuous vector representations in a high-dimensional space. These embeddings encode syntactic and semantic properties such that tokens with related meanings occupy nearby regions in the embedding space~\cite{mikolov2013efficient}. Positional encodings are further incorporated to preserve word order information, enabling the model to reason over sequences of words. This embedding-based representation serves as the foundation upon which all subsequent learning stages operate.

The training of LLMs typically follows a two-stage paradigm: \emph{pretraining} followed by \emph{fine-tuning}. Pretraining is a large-scale, computation-intensive phase in which the model is exposed to massive amounts of unlabeled or weakly labeled text using self-supervised objectives, most commonly next-token prediction~\cite{mckinzie2024mm1}. Through this process, the model acquires general linguistic competence, including grammar, semantics, and broad world knowledge \cite{biderman2023pythia,touvron2023llama}. The resulting pretrained model functions as a general-purpose language representation but is not inherently aligned with specific user intents or downstream tasks~\cite{du2024stacking}.

Fine-tuning constitutes the specialization stage, where the pretrained model is adapted to particular tasks, domains, or interaction styles using comparatively small, curated datasets~\cite{dettmers2023qlora}. This adaptation typically relies on supervised learning or preference-based objectives, adjusting the pretrained weights to improve task relevance, instruction following, or safety properties \cite{bianchi2023safety,groeneveld2024olmo}. Recent studies highlight that high-quality fine-tuning data can yield substantial performance gains even with limited dataset sizes, emphasizing data efficiency over scale \cite{zhou2023lima}. Moreover, emerging analyses suggest that pretraining and fine-tuning are not independent processes but form a coupled system, where the structure of pretrained representations strongly influences fine-tuning efficiency and outcomes \cite{sundredze2025amuro}.

From a systems perspective, this two-stage training paradigm has profound implications for data, memory, and compute efficiency. While pretraining dominates overall computational cost, fine-tuning increasingly serves as the primary mechanism for rapid adaptation and deployment. Consequently, optimizing fine-tuning strategies, through parameter-efficient updates, data selection, and memory-aware training techniques, has become a central research focus, motivating the methodologies discussed in the following subsection.

\subsection{Methodologies of Fine‑Tuning}\label{finetune}

\begin{figure}[t]
  \centering
  \input{finetuningLLM_tikz.tex}
  \caption{Types of fine-tuning methods.}
  \label{fig:finetuning-llms}
\end{figure}
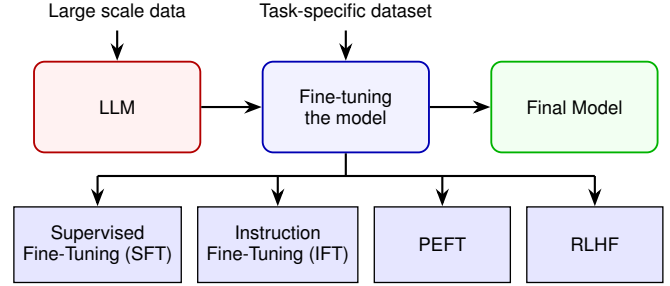

Fine‑tuning LLMs can be broadly categorized into several complementary approaches, as shown in Figure~\ref {fig:finetuning-llms}. These approaches exhibit distinct trade-offs in terms of data requirements, memory footprint, computational cost, and downstream performance.

\subsubsection{Parameter‑Efficient Fine‑Tuning (PEFT)}

PEFT methods seek to adapt pretrained LLMs by modifying only a small fraction of their parameters, relying on the hypothesis that model adaptation has a low intrinsic dimension~\cite{aghajanyan2021intrinsic,ding2022delta}. This approach drastically reduces memory consumption and permits fine‑tuning on resource‑constrained hardware. Classic PEFT methods include inserting small adapter modules between layers (“Adapters”)~\cite{karimi2021compacter}, optimizing continuous soft prompts or prefix embeddings (“Prefix Tuning”)~\cite{li2021prefix}, or updating only bias terms rather than full weight matrices (“BitFit”)~\cite{ben2022bitfit}.

Among these, one of the most influential is LoRA (Low‑Rank Adaptation)~\cite{hu2021lora}, which freezes all original weights and learns only a pair of low-rank adapter matrices per layer. While PEFT typically yields substantial gains in memory efficiency and training speed, the heavy parameter reduction or structural constraints may reduce expressiveness. To address this, recent variants expand PEFT’s flexibility via dynamic-rank adapters and split structures~\cite{lin2024splitlora,kopiczko2023vera}, multi-branch or hybrid parameterization~\cite{mao2022unipelt,wu2024advancing}, and new optimizer-aware techniques such as GaLore~\cite{zhao2024galore} which further reduce memory overhead by projecting gradients into low-rank subspaces.

\subsubsection{Supervised Fine-Tuning (SFT)}

In SFT, the model is trained on labeled datasets of input–output examples specific to a target task. During training, model parameters are updated to minimize the discrepancy (typically cross-entropy loss) between the predicted and true outputs, enabling the model to generalize to unseen data. This approach remains the most straightforward adaptation method and acts as a critical ``recipe" for unlocking capabilities in smaller models~\cite{pareja2024unveiling}.

SFT is widely used as the primary stage for tasks like summarization or translation, yet its success is highly sensitive to the training signal. Extensive experiments reveal that data composition~\cite{dong2024abilities}, layer-wise dynamics~\cite{harada2025massive}, and fine-grained token selection~\cite{pang2025token} are decisive factors in shaping alignment quality. Furthermore, while standard SFT is effective, relying solely on cross-entropy minimization can lead to overfitting or a collapse in generation diversity. To mitigate this, recent approaches propose entropic or diversity-preserving objectives that maintain the richness of the model's output distribution~\cite{li2024entropic,li2024preserving}.
 
\subsubsection{Instruction Fine-Tuning (IFT)}

IFT, often referred to as “instruction tuning,” extends SFT by utilizing datasets composed of natural-language instructions paired with corresponding responses~\cite{mishra2022cross}. The primary goal is to teach the LLM to follow human directives and perform a variety of tasks without the need for retraining separate models for each specific objective. 

This paradigm fundamentally shifts the model's capability from simple pattern matching to task interpretation, enabling robust \textit{zero-shot generalization} across unseen tasks~\cite{wei2021finetuned,sanh2021multitask}. By exposing the model to a diverse range of instructional templates during training, IFT broadens applicability and usability. Furthermore, when combined with PEFT strategies, IFT becomes a highly efficient mechanism for aligning model behavior with human intent without the prohibitive cost of full-parameter updates.

\subsubsection{Reinforcement Learning from Human Feedback (RLHF)}

For tasks where desired output quality is subjective, such as helpfulness, safety, or nuance, standard supervised signals are often insufficient. To address this, RLHF incorporates human judgments directly into the training loop~\cite{christiano2017deep,ziegler2019fine}. In the standard pipeline, human evaluators rank model outputs to train a separate reward model, which then acts as a proxy to guide the LLM via reinforcement learning (RL) algorithms like Proximal Policy Optimization (PPO)~\cite{schulman2017proximal}.

This methodology was pivotal in developing modern instruction-following models, aligning raw probabilistic predictions with human values~\cite{ouyang2022training}. However, the complexity of managing separate reward models has led to the emergence of direct alignment algorithms. Methods such as Direct Preference Optimization (DPO)~\cite{rafailov2023direct} and its variants~\cite{song2024preference,xu2024contrastive} optimize the policy directly on preference data without an explicit reward modeling step. Despite these advances, preference learning remains challenging, facing open problems regarding reward hacking, data efficiency, and the fundamental limitations of using human feedback as a gold standard~\cite{casper2023_rlhf_limits}.

\subsubsection{Hybrid and Combined Strategies}

In practice, reliance on a single tuning paradigm often involves trade-offs between performance and computational cost. To address this, modern workflows increasingly adopt hybrid strategies that combine the strengths of unitary methods to mitigate their individual weaknesses~\cite{qi2025hybridunitaryfinetuninglarge}. A common pattern involves a sequential pipeline: a base model first undergoes broad SFT or IFT to establish instruction-following capabilities, followed by parameter-efficient adaptation (e.g., LoRA) to specialize for specific domains like industrial diagnostics or multi-task classification~\cite{pang2024hybrid,beiranvand2025hybrid}. 

Beyond simple sequencing, recent research explores structural hybrids, such as optimizing adaptation strategies ``bottom-up" across layers to extend the potential of efficient tuning~\cite{guetal2024bottom}. Such strategies are particularly appealing for deploying LLMs on edge devices, where they balance the high alignment quality required for safety with the strict memory and compute constraints of local hardware.

\section{The Foundation: Data Efficiency (The ``What" to Train On)}\label{data}

\begin{figure}[t]
    \centering
    \resizebox{\columnwidth}{!}{%
    \footnotesize
    \begin{forest}
      for tree={
        grow'=0, 
        draw,
        align=center,
        node options={rounded corners, inner sep=3pt},
        edge={thick},
        l sep=0.7cm, 
        s sep=0.4cm, 
        font=\sffamily
      }
      [\textbf{Data Efficiency} \\ \textit{(Selection Strategies)}, 
        [\textbf{Foundational Metrics} \\ \textit{(Early Pruning)}, fill=gray!10
            [LIMA \cite{zhou2023lima} \\ \scriptsize{Few-shot Alignment}]
            [GraNd / EL2N \cite{paul2021deep} \\ \scriptsize{Gradient/Error Norm}]
        ]
        [\textbf{Learning Dynamics} \\ \textit{(Proxy Models)}, fill=blue!10
            [SmallToLarge (S2L) \cite{yang2024smalltolarge} \\ \scriptsize{Loss Trajectories}]
            [STAFF \cite{zhang2025staff} \\ \scriptsize{Speculative Selection}]
        ]
        [\textbf{Gradient Influence}, fill=red!10
            [LESS \cite{xia2024less} \\ \scriptsize{Adam Influence}]
            [GREATS \cite{wang2024greats} \\ \scriptsize{Online Taylor Exp.}]
            [Dynamic Influence \cite{wang2025dynamic} \\ \scriptsize{Normalized \& Dynamic}]
        ]
        [\textbf{Multidimensional} \\ \textit{(Quality \& Curriculum)}, fill=green!10
            [BIDS \cite{dai2025bids} \\ \scriptsize{Balanced Capabilities}]
            [DART \cite{tong2024dart} \\ \scriptsize{Difficulty-Aware}]
            [IFD \cite{li2024quantityqualityboostingllm} \\ \scriptsize{Instruction Gain}]
            [AlpaGasus/MoDS \cite{chen2024alpagasus} \\ \scriptsize{Teacher Scoring}]
        ]
      ]
    \end{forest}
}
    \caption{\textbf{Taxonomy of Data Efficiency Methods.} We categorize selection strategies into four branches: early pruning based on static metrics, dynamics-based selection using proxy models, mathematical gradient influence on the target model, and multidimensional approaches focusing on difficulty, balance, and quality.}
    \label{fig:data_efficiency_taxonomy}
\end{figure}
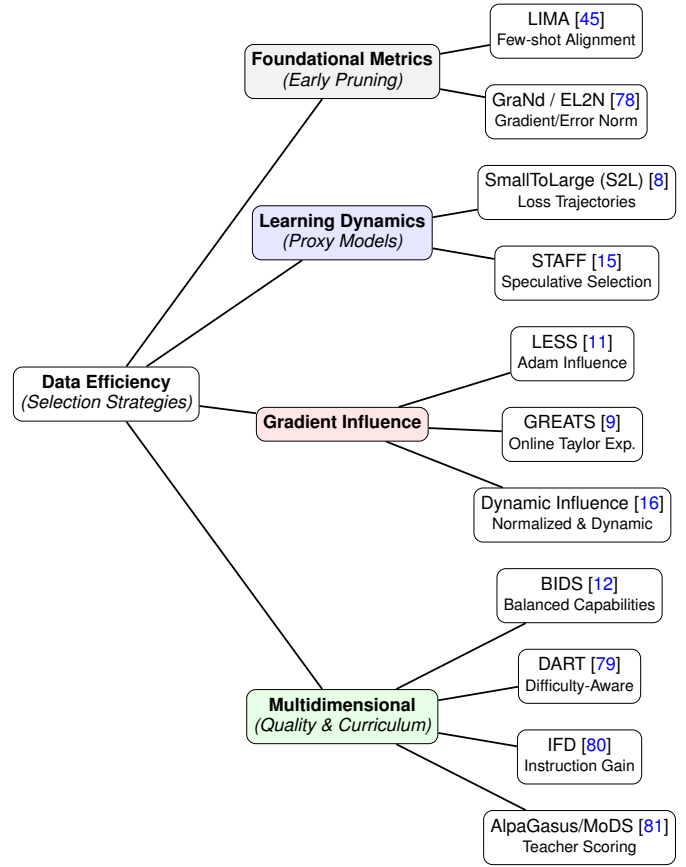

Modern fine-tuning methods such as SFT and RLHF are often constrained by the quantity, quality, and redundancy of the training data. Although contemporary LLMs are trained on massive datasets, not all examples contribute equally to performance. Data-efficient selection techniques aim to maximize learning from fewer, high-impact samples, reducing computational costs and improving generalization.

To ground these empirical methods, recent studies provide a theoretical framework for data pruning.
Du et al. argue that data pruning can be decomposed into two distinct components: the \textit{data representation} and the \textit{selection algorithm} \cite{du2025disentangling}. A key finding is that the choice of representation is often more critical than the selection algorithm itself. Theoretical analysis reveals that gradients are generally the most effective representation because they reflect the distance to the decision boundary and encode label information. In contrast, hidden states often lack this discriminative power \cite{du2025disentangling}. Furthermore, the study highlights that selection objectives are context-dependent. Relevance-based objectives excel under distribution shifts, while difficulty-based objectives are superior in high-budget settings.

In another comprehensive empirical study, Liu et al. \cite{liu2024what} systematically analyze the impact of three core data dimensions: Quality, Complexity, and Diversity (QCD). Their findings challenge the common intuition that diversity is paramount. They demonstrate that for alignment tasks, \textit{Complexity} (e.g., the reasoning depth and length of the response) is often the single most critical factor for performance, followed by quality. Diversity serves mainly to prevent overfitting but has diminishing returns. This suggests that efficient selection strategies should prioritize complex, high-reasoning examples over merely accumulating a diverse set of simple instructions \cite{liu2024what}.

\subsection{Foundational Metrics for Early Pruning}

Before the advent of complex influence functions for LLMs, foundational work established that data importance could be identified early in the training process. This allows for the removal of redundant examples without waiting for full model convergence.

\paragraph{The Superficial Alignment Hypothesis (LIMA)}
Challenging the assumption that alignment requires massive datasets, LIMA \cite{zhou2023lima} demonstrates that a strong pre-trained model can achieve competitive performance using only 1,000 carefully curated examples. This ``Less Is More for Alignment" principle serves as the foundational motivation for data efficiency, proving that data quality and diversity often outweigh sheer quantity.

\paragraph{Gradient Norm and Error $L_2$-Norm}
Paul et al. proposed methods to prune datasets by identifying significant examples after only a few epochs \cite{paul2021deep}. They introduced two key metrics to quantify this significance for a data point $z$ consisting of input $x$ and label $y$. First, the Gradient Normed Score (GraNd) measures the expected magnitude of the loss gradient vector. It bounds the potential influence of a training example on reducing the loss of any other example in a single optimization step:

\begin{equation}
    \mathrm{GraNd}_t(z) = \mathbb{E}_{\theta_t}\,\bigl\lVert \nabla_{\theta_t}\, \mathcal{L}(z; \theta_t) \bigr\rVert_2.
\end{equation}


Here, $\theta_t$ represents the model parameters at epoch $t$, and the expectation $\mathbb{E}$ accounts for stochasticity (e.g., dropout). Second, they introduced the Error $L_2$-Norm Score (EL2N). This is a practical and empirically often superior approximation defined as the expected $L_2$-norm of the error vector between the predicted probabilities $p(\theta_{t}, x)$ and the one-hot label $y$:

\begin{equation}
\text{EL2N}_t(z) = \mathbb{E}||\,p(\theta_{t}, x) - y\,||_{2}.
\end{equation}


This work demonstrated that a large fraction of data can be discarded early in the training process without sacrificing final model performance \cite{paul2021deep}.

\subsection{Leveraging Learning Dynamics and Proxy Models}

A promising approach to increasing data efficiency involves analyzing learning dynamics during the training process. Instead of calculating expensive gradients on the target model, these methods leverage the consistency of training dynamics between small and large models.

\paragraph{Trajectory-Based Selection (SmallToLarge)}

A prominent method in this field is ``SmallToLarge'' (S2L) \cite{yang2024smalltolarge}. Instead of merely filtering data statically, this method examines how the error of a model for individual examples evolves over time. Let $\theta_{\text{proxy}}^{(t)}$ be the parameter set of a small proxy model at a training time $t$. The loss trajectory for a data point $z$ is recorded as a vector $\mathbf{T}_z^{\text{proxy}}$ containing error values $\mathcal{L}$ at timestamps $t_1$ to $t_T$:


\begin{equation}
    \mathbf{T}_z^{\text{proxy}} = [\mathcal{L}(z; \theta_{\text{proxy}}^{(t_1)}), \dots, \mathcal{L}(z; \theta_{\text{proxy}}^{(t_T)})].
\end{equation}


The theoretical foundation relies on the Hessian matrix describing the curvature of the loss landscape. Assuming that this curvature is bounded, it can be shown that examples with similar loss trajectories also generate similar gradients. If two examples $z_i$ and $z_j$ exhibit similar error curves on the small model, the difference of their gradients in the large target model $\theta_{\text{target}}$ is bounded by an upper limit $\Delta$:


\begin{equation}
    \|\nabla \mathcal{L}(z_i; \theta_{\text{target}}) - \nabla \mathcal{L}(z_j; \theta_{\text{target}})\| \leq \Delta.
\end{equation}


This insight implies that data points with a similar error history influence the model in an almost identical direction during training and are thus redundant. The S2L method clusters these trajectories and samples a subset. Empirical results are significant. On the MathInstruct dataset \cite{yue2023mammoth}, S2L matches full-dataset performance while using only 11\% of the original examples. Notably, training on a selected subset of 50,000 examples improves the accuracy of the Phi-2 model on the challenging MATH benchmark by 16.6\% \cite{yang2024smalltolarge}.

\paragraph{Speculative Coreset Selection (STAFF)}

Similar to S2L, the STAFF method leverages the efficiency of smaller models but adopts the concept of speculative execution from computer architecture \cite{zhang2025staff}. This approach addresses the high computational overhead of calculating influence scores directly on the target model. STAFF employs a two-stage process using a smaller proxy model $\theta_{\text{proxy}}$ and the target model $\theta_{\text{target}}$ from the same family.


First, in the \textit{Speculative Score Calculation} stage, STAFF utilizes the efficient small model to estimate the difficulty of each sample $z$. It employs the Effort Score, defined as the $L_2$-norm of the gradient of the loss with respect to the proxy parameters:


\begin{equation}
    S_{z}^{\text{proxy}} = ||\nabla_{\theta_{\text{proxy}}}\mathcal{L}(z; \theta_{\text{proxy}})||_{2}.
\end{equation}


This score reflects the magnitude of parameter updates required to fit the sample. However, to account for distributional differences, STAFF introduces a \textit{Verification} stage. The dataset is stratified into regions based on $S^{\text{proxy}}$. For each region $i$, STAFF draws a small stratified verification subset $\mathcal{B}_{i}^{*}\subseteq \mathcal{B}_i$ (i.e., samples randomly selected from region $i$ as defined by the proxy-score bins) and evaluates it on the target model to compute a verification factor $\mathcal{V}_i$:


\begin{equation}
    \mathcal{V}_{i} = \frac{\sum_{z\in \mathcal{B}_{i}^{*}}S_{z}^{\text{target}}}{\sum_{z\in \mathcal{B}_{i}^{*}}S_{z}^{\text{proxy}}}.
\end{equation}


Here, $\mathcal{V}_{i} > 1$ indicates that a region is more critical to the target model than predicted by the proxy. This factor is subsequently used to dynamically adjust the selection budget for each region, ensuring that sampling prioritizes data that is specifically important to the target architecture while maintaining diversity. Empirical evaluations demonstrate that STAFF can reduce selection overhead by up to 70.5\% compared to standard methods while improving fine-tuning performance by up to 54.3\% \cite{zhang2025staff}.

\subsection{Targeted Selection via Gradient Influence}\label{gradient}

While proxy models offer scalability, gradient-based methods provide a more granular and mathematically rigorous assessment of how specific data points impact the target task. These methods typically rely on influence functions or gradient projections.

\paragraph{Low-Rank Gradient Similarity (LESS)}

Xia et al. introduce LESS (Low-rank Gradient Similarity Search) to select data based on gradient similarity \cite{xia2024less}. The goal is to quantify the influence of a training example $z_{tr}$ on a validation set represented by $z_{val}$. To define an influence compatible with the Adam optimizer, LESS proposes the \textit{Adam Influence} score. This score is accumulated over a trajectory of checkpoints $t=1 \dots T$:


\begin{equation}
    \text{Inf}_{\text{Adam}}(z_{tr}, z_{val}) = \sum_{t=1}^{T} \eta_t \cos(\nabla \mathcal{L}(z_{val}; \theta_t), \Gamma(z_{tr}; \theta_t)).
\end{equation}


Here, $\eta_t$ is the learning rate (LR) and $\Gamma(z_{tr}; \theta_t)$ represents the preconditioned update vector used by Adam (incorporating momentum and variance) rather than the raw gradient. To make this computation tractable for billions of parameters, LESS utilizes random projections: high-dimensional (LoRA) gradient/update vectors are projected into a $d$-dimensional space using a random matrix $\Pi^{\top}$. In practice, LESS uses a relatively large projection dimension (default $d=8192$), and ablations consider $d \in \{1024, 2048, 4096, 8192\}$ to study the fidelity--memory trade-off. By selecting the top examples with this method, LESS demonstrates that training on a 5\% subset can outperform training on the full dataset \cite{xia2024less}.


\paragraph{Online Selection (GREATS)}
While LESS focuses on static dataset pruning, Wu et al. introduce GREATS (GREedy Approximation Taylor Selection) to address the dynamic nature of learning via online batch selection \cite{wang2024greats}. GREATS formulates the selection of a training batch subset $\mathcal{S}$ as a set function optimization problem, where the goal is to maximize the utility $U^{(t)}(\mathcal{S})$, defined as the reduction in validation loss after a gradient step on $\mathcal{S}$.


Since exact evaluation of this utility is computationally prohibitive, GREATS approximates the marginal gain of adding a candidate sample $z_{new}$ to an already selected subset $\mathcal{S}_t$ using a second-order Taylor expansion:

\begin{equation}
\begin{split}
    U^{(t)}(z_{new} | \mathcal{S}_t) &\approx \underbrace{\eta_t g(z_{new}) \cdot g(z_{val})}_{\text{Alignment}} \\
    &\quad - \underbrace{\eta_t^2 g(z_{new}) H(z_{val}) \sum_{z \in \mathcal{S}_t} g(z)}_{\text{Redundancy Correction}}, \\
    \text{where }
g_t(z) &:= \nabla_{\theta}\mathcal{L}(z;\theta_t), 
H_t(z_{\text{val}}) := \nabla_{\theta}^2 \mathcal{L}(z_{\text{val}};\theta_t).
\end{split}
\end{equation}

Here, $g_t(z)$ denotes the gradient vector $\nabla \mathcal{L}(z; \theta_t)$ and $H(z_{val})$ represents the Hessian matrix on the validation data. This approximation reveals two key components: an \textit{Alignment} term (similar to TracIN~\cite{pruthi2020estimating}) that measures how well a sample reduces validation loss, and a \textit{Redundancy Correction} term that penalizes samples with gradients similar to those already selected ($z \in \mathcal{S}_t$), thus enforcing diversity. To make this computationally feasible during training, GREATS introduces the \textit{Ghost Inner-Product} technique, avoiding the instantiation of model-sized gradient vectors. Empirical results show that GREATS significantly accelerates convergence and improves generalization even with very small validation sets (e.g., $N_{val} \le 16$) \cite{wang2024greats}.

\paragraph{Dynamic Gradient-Based Selection}
\label{sec:gradient-section}
Wang et al. further refine this area by identifying and addressing two critical limitations of traditional one-step gradient methods: selection length bias and decreasing long-term effectiveness \cite{wang2025dynamic}. Through theoretical analysis, they reveal that the gradient norm of a training sample tends to decrease as the sequence length $N$ increases ($||\nabla\mathcal{L}|| \sim O(N^{-q})$). This causes standard influence approximations to erroneously favor shorter, less informative sequences. To mitigate this, they propose a normalized influence score. When using the Adam optimizer, the influence of a training sample $z_{tr}$ on a validation sample $z_{val}$ at step $t$ is approximated as

\begin{equation}
    \tilde{\mathcal{I}}^t(z_{tr}, z_{val}) \approx \left\langle \frac{\nabla\mathcal{L}(z_{val}; \theta_t)}{||\nabla\mathcal{L}(z_{val}; \theta_t)||}, \frac{\Gamma(z_{tr}; \theta_t)}{||\Gamma(z_{tr}; \theta_t)||} \right\rangle,
\end{equation}
where $\Gamma$ represents the Adam-based parameter update vector and $\langle \cdot, \cdot \rangle$ denotes the standard Euclidean inner product in the model's parameter space, $\mathbb{R}^P$, where $P$ is the number of trainable model parameters. Both the parameter update vector $\Gamma$ and the gradient vector $\nabla\mathcal{L}$ are implicitly vectorized (flattened) when computing the inner product. Similar to LESS, to make the computation tractable for models with billions of parameters, these high-dimensional vectors are then projected into a lower-dimensional space $\mathbb{R}^d$ (e.g., $d=8192$ dimensions) using random projection before computing the similarity score.  Furthermore, to counter the diminishing correlation between initial influence scores and actual loss reduction over time, they introduce a dynamic selection framework. Instead of a static coreset, the influence scores are periodically recomputed (e.g., every epoch), and the data coreset is dynamically updated to reflect the model's evolving training state. Empirical results demonstrate that this dynamic, normalized approach consistently outperforms static methods like LESS on general benchmarks \cite{wang2025dynamic}.


\subsection{Multidimensional Selection: Balance, Difficulty and Curriculum}

Recent advancements emphasize that mathematical influence alone is insufficient. To construct optimal training sets, one must also consider the balance between diverse capabilities and the intrinsic difficulty of the examples.

\paragraph{Balancing Capabilities (BIDS)}

Dai et al. identify an inherent bias in influence-based methods where certain tasks naturally exhibit higher influence magnitudes than others \cite{dai2025bids}. This leads to naive algorithms oversampling high-influence tasks. To mitigate this, BIDS (Balanced and Influential Data Selection) introduces a normalization framework. Let $\mathcal{I}_{ij}$ denote the influence of a training sample $z_i$ on a validation sample $z_j$. This score approximates the expected reduction in loss on $z_j$ if the model is trained on $z_i$, typically calculated as the inner product of their gradients:
\begin{equation}
    \mathcal{I}_{ij} \approx \nabla \mathcal{L}(z_i; \theta)^\top \nabla \mathcal{L}(z_j; \theta).
\end{equation}
However, since the magnitude of gradients varies across tasks, BIDS normalizes this score using the mean $\mu_j$ and standard deviation $\sigma_j$ calculated across all training samples for the $j$-th validation instance:


\begin{equation}
    \tilde{\mathcal{I}}_{ij} = \frac{\mathcal{I}_{ij} - \mu_j}{\sigma_j}.
\end{equation}


Following normalization, BIDS employs an iterative greedy selection strategy. It selects the candidate $z^*$ from the pool $\mathcal{D}_{pool}$ that maximizes the marginal gain for the \textit{minimum} capability (weakest validation task) in the current set $\mathcal{S}$:


\begin{equation}
    z^* = \arg\max_{z \in \mathcal{D}_{pool} \setminus \mathcal{S}} \left( \min_{j \in \mathcal{D}_{val}} \left( \sum_{s \in \mathcal{S} \cup \{z\}} \tilde{\mathcal{I}}_{sj} \right) \right).
\end{equation}


This objective forces the algorithm to prioritize examples that strengthen the weakest link in the model's capability profile \cite{dai2025bids}.

\paragraph{Difficulty-Aware Selection (DART)}

In complex reasoning domains like mathematics, the intrinsic difficulty of examples plays a crucial role. Tong et al. highlight that standard rejection tuning often fails to generate correct responses for difficult queries \cite{tong2024dart}. DART (Difficulty-Aware Rejection Tuning) introduces a strategy based on the \textit{Failure Rate} ($r_{\text{fail}}$). Concretely, for each query, a fixed number of \emph{raw candidate responses} is sampled to estimate its difficulty. Let $n$ denote this per-query sample count used for difficulty evaluation, and let $n_{\text{correct}}$ be the number of answer-correct responses among these $n$ candidates. The failure rate is defined as:

\begin{equation}
    r_{\text{fail}} = 1 - \frac{n_{\text{correct}}}{n}.
\end{equation}

DART applies a ``Prop2Diff" strategy where the sampling budget $K$ is allocated proportional to difficulty ($K \propto r_{\text{fail}}$). This ensures that a significantly higher sampling budget is allocated to difficult queries to increase the probability of capturing correct reasoning paths \cite{tong2024dart}.


\paragraph{Linear Indicator Mining}
To avoid the computational cost of influence functions or large teacher models, Cao et al. propose a lightweight selection method based on linear indicators \cite{cao2024instruction}. They identify that simple statistical features such as input length, output length, and perplexity often correlate strongly with data quality. By fitting a linear regression model to predict the ``quality" (defined by downstream performance) based on these indicators, they can mine high-quality instruction data from massive corpora efficiently. This approach suggests that complex semantic scoring is not always necessary if robust statistical proxies are available.

\paragraph{Instruction-Following Difficulty (IFD)}
Building on the insights from LIMA \cite{zhou2023lima} regarding data quality, Li et al. introduce a self-guided selection methodology termed Instruction-Following Difficulty (IFD) \cite{li2024quantityqualityboostingllm}. Operating on the premise that models should focus on samples where instructions provide significant information gain, IFD identifies ``cherry" samples where the instruction $x$ is critical for generating the correct response $y$. The metric is defined as the ratio between the model’s Conditioned Answer Score (loss given instruction) and its Direct Answer Score (loss without instruction):
\begin{equation}
    r_{\text{IFD}}(z) = \frac{\mathcal{L}(y \mid x; \theta)}{\mathcal{L}(y \mid  \theta)}.
\end{equation}
This ratio effectively filters out trivial examples that the model can already resolve via pre-training knowledge (where $\mathcal{L}(y)$ is low). The approach employs a three-phase pipeline: learning from a brief experience to estimate difficulty, scoring the full dataset, and retraining on the selected subset. Empirical evaluations demonstrate that fine-tuning on merely 5\% of data selected via IFD allows the model to surpass baselines trained on the full dataset \cite{li2024quantityqualityboostingllm}.

\paragraph{Model-Based Quality Scoring (AlpaGasus \& MoDS)}
A pragmatic alternative to calculating mathematical influence is leveraging strong teacher models (e.g., GPT-4) to explicitly score data quality. Chen et al. introduce AlpaGasus \cite{chen2024alpagasus}, which filters the Alpaca dataset by using a powerful LLM to score each (instruction, input, output) tuple. By keeping only examples with high scores, AlpaGasus matches the performance of the original model with significantly fewer data points, demonstrating that ``stronger" models can effectively distill data for ``weaker" ones. Similarly, MoDS (Model-oriented Data Selection) \cite{du2024mods} adopts a multi-perspective approach. It evaluates data based on three criteria: quality (scored by an LLM), coverage (to ensure diversity), and necessity (preventing redundancy). This holistic filtering reduces the dataset size while maintaining high instruction-following capability.

\paragraph{Curriculum and Compute-Awareness}

Extending the concept of difficulty, curriculum-based approaches argue that the usefulness of an example changes as the model's competence evolves \cite{yincompute}. Simpler samples accelerate learning in early stages, while harder examples become crucial later. This complements difficulty-based sampling by shifting the focus to \textit{when} a sample is beneficial. In this context, we also recall the approach by Wang et al. \cite{wang2025dynamic} (previously discussed in subsection \ref{gradient}), which aligns with the principle of evolving data utility. Additionally, Yu et al. introduce a dynamic perspective driven by compute constraints \cite{yu2025llm}. Their method adaptively determines which data to use based on the marginal utility under a limited token budget. This formalizes adaptive pacing through a bi-level optimization framework that learns per-sample utility weights \cite{yu2025llm}.

\subsection{Synthesis: From Static Filtering to Dynamic Marginal Utility}


The reviewed literature reveals a fundamental shift in data efficiency. The field is moving from static noise filtering found in methods like LIMA or AlpaGasus to maximizing the \textit{Marginal Utility per Token}. This is achieved via metrics based on gradients like LESS and GREATS or multidimensional metrics like BIDS and DART. However, this survey identifies a critical tension within the optimization trilemma. While sophisticated gradient methods offer theoretical \textit{optimality} by identifying the most influential data, they often sacrifice \textit{feasibility} due to their exorbitant computational and memory costs.

This creates what we term the \textit{Static to Dynamic Gap}. Current leading methods like LESS remain predominantly static. Reevaluating data influence during training is essential to capture evolving learning dynamics. However, this requires storing massive gradient histories or performing frequent backpropagation passes. Consequently, ``perfect" data selection is often rendered impractical by the very hardware constraints it aims to mitigate. This creates a direct trade-off: using a more complex, memory-intensive data selector (e.g., gradient-based) might yield a higher-quality model but consumes a significant portion of the VRAM that could otherwise be used to increase the context window or batch size (memory efficiency) on edge-grade GPUs. True data efficiency cannot be solved by selection algorithms alone.
A practical bridge is to approximate dynamic influence with memory-efficient proxies (e.g., CoLM-style coresets or low-rank sketches), but this introduces a stability trade-off: the selector observes a proxy update $\hat g_t(z)=g_t(z)+\varepsilon_t(z)$ (and often a stale estimate if refreshed every step), which can perturb influence rankings and lag behind shifting decision boundaries. In a closed-loop pipeline, such noise/lag can compound into oscillatory sampling signals for the governor (Sec.~\ref{compute}).


We view this hybridization as a three-way trade-off between \textit{fidelity} (smaller $\|\epsilon_t\|$), \textit{responsiveness} (smaller $K$), and \textit{stability}. To overcome these hurdles, we propose a research roadmap with the following technical milestones:
\begin{itemize}
    \item \textbf{Drift-aware Refresh Schedules:} Using gradient distribution divergence to trigger re-calibration.
    \item \textbf{Hybrid Proxy-Verify Pipelines:} Sparse target-model verification (STAFF-style) to calibrate proxy scores and eliminate lag.
    \item \textbf{Damped Governor Updates:} Using EMA, hysteresis, or trust-region limits on sampling-weight changes to prevent ``chatter''.
    \item \textbf{Incremental Influence Updates:} Developing methods to update scores instead of full recomputation to optimize net marginal utility per token.
\end{itemize}
While data selection reduces the total number of tokens, its ultimate efficiency is capped by the memory-compute trade-offs discussed in the following sections. To navigate these limits, we later formalize this dynamic data selection process as a primary component of the state vector $S_t$ within the integrated compute governor framework (Section \ref{compute}). This allows the system to use feedback on loss dynamics and gradient variance to adaptively balance sample quality against available hardware resources.

\section{The Constraint: Memory Efficiency (The ``How" to Fit It)}\label{memory}

In practice, scaling LLMs encounters the hard limit of GPU memory (VRAM) long before it hits the boundaries of computational throughput. To understand the solutions, we must first decompose the memory consumption $M_{total}$ of training a model with parameters $\theta$:
\begin{equation}
    M_{total} \approx \underbrace{M_{\theta}}_{\text{Weights}} + \underbrace{M_{\mathcal{O}}}_{\text{Optimizer States}} + \underbrace{M_{\mathcal{A}}}_{\text{Activations}}.
\end{equation}
While PEFT methods like LoRA~\cite{hu2021lora} or AdaLoRa~\cite{zhang2023adalora} successfully mitigate $M_{\mathcal{O}}$ by reducing trainable parameters, they often fail to address $M_{\mathcal{A}}$, the activation memory required for backpropagation, which scales linearly with sequence length and batch size. Even with hardware-aware optimizations like FlashAttention~\cite{dao2022flashattention}, the bottleneck persists for long-context tasks.

Consequently, recent research has moved beyond simple parameter reduction to attack each term of this equation directly~\cite{lin2025enhancing, pudipeddi2020training}. This section surveys four strategic levers that redefine the ``how" of training: \textit{Data-Centric Selection} (reducing input dimensions), \textit{Block-wise Optimization} (fragmenting optimizer states), \textit{Gradient-Free Approximation} (eliminating activation storage), and Quantization-Centric Approach.

\begin{figure}[htbp]
    \centering
    \resizebox{\linewidth}{!}{
    \begin{tikzpicture}[
        node distance=0.4cm and 0.8cm, 
        font=\sffamily\footnotesize,
        memblock/.style={
            rectangle, draw=black!60, fill=white, thick,
            align=center, minimum height=1.6cm, text width=3.2cm,
            inner sep=3pt
        },
        stratblock/.style={
            rectangle, rounded corners, align=center, 
            minimum height=1.4cm, text width=3.4cm, 
            fill opacity=0.2, text opacity=1, inner sep=3pt
        },
        arrow_style/.style={->, >=Stealth, thick}
    ]

    
    \node[memblock, fill=red!5, anchor=north west, xshift=0.6cm] (weights_mem) {
        \textbf{Weights ($M_{\theta}$)}\\
        \scriptsize FP32/FP16 Parameters
    };

    \node[memblock, fill=green!5, below=of weights_mem] (opt_mem) {
        \textbf{Optimizer ($M_{\mathcal{O}}$)}\\
        \scriptsize States (Mom., Var.)\\ \scriptsize (e.g., Adam: $2\times$ params)
    };

    \node[memblock, fill=orange!5, below=of opt_mem, minimum height=2.5cm] (act_mem) {
        \textbf{Activations ($M_{\mathcal{A}}$)}\\
        \scriptsize Stored for Backprop\\ \scriptsize (Scales w/ batch \& seq)
    };

    
    \draw[line width=2pt, gray!70] 
        ([xshift=-6pt]weights_mem.north west) -- ([xshift=-6pt]weights_mem.south west) 
        node[midway, left=2pt, rotate=90, font=\scriptsize\bfseries, text=gray!90!black] {Static};
        
    \draw[line width=2pt, gray!70] 
        ([xshift=-6pt]opt_mem.north west) -- ([xshift=-6pt]act_mem.south west) 
        node[midway, left=2pt, rotate=90, font=\scriptsize\bfseries, text=gray!90!black] {Dynamic};


    \node[stratblock, fill=red, right=of weights_mem] (strat_quant) {
        \textbf{Quantization-Centric}\\
        \scriptsize Reduce precision (INT8/4)\\
        \textit{DQT~\cite{zhao2024direct}, PEQA~\cite{kim2023memoryefficientfinetuningcompressedlarge}, Q-LoRA~\cite{dettmers2023qlora}}\\[1ex]
        {\tiny\bfseries\textcolor{blue!80!black}{[Train \& Inference]}}
    };

    \node[stratblock, fill=green!80!black, right=of opt_mem] (strat_block) {
        \textbf{Optimizer-Centric}\\
        \scriptsize Block-wise updates\\
        \textit{HiFT \cite{liu2024hift}, BAdam \cite{luo2024badam}}\\[1ex]
        {\tiny\bfseries\textcolor{blue!80!black}{[Training Only]}}
    };

    \node[stratblock, fill=orange, right=of act_mem.north east, anchor=north west, yshift=+0.2cm, minimum height=1.2cm] (strat_data) {
        \textbf{Data-Centric}\\
        \scriptsize Reduce Input Size\\
        \textit{CoLM~\cite{nguyen2025minibatch}, Addax~\cite{li2024addax}, QLESS~\cite{ananta2025qlessquantizedapproachdata}}\\[0.5ex]
        {\tiny\bfseries\textcolor{blue!80!black}{[Training Only]}}
    };

    \node[stratblock, fill=purple, right=of act_mem.south east, anchor=south west, yshift=-0.6cm, minimum height=1.2cm] (strat_radical) {
        \textbf{Radical (ZO)}\\
        \scriptsize Activation-free est.\\
        \textit{SubZero~\cite{yu2024subzero}, LOZO~\cite{chen2024enhancing}}\\[0.5ex]
        {\tiny\bfseries\textcolor{blue!80!black}{[Training Only]}}
    };

    
    \draw[arrow_style, color=red!80] (strat_quant.west) -- (weights_mem.east);
    \draw[arrow_style, color=red!80] (strat_quant.west) -- (opt_mem.east);
    \draw[arrow_style, color=red!80] (strat_quant.west) -- (act_mem.east);
    
    \draw[arrow_style, color=green!60!black] (strat_block.west) -- (opt_mem.east);
    
    \draw[arrow_style, color=orange!80!black] (strat_data.west) -- (act_mem.east);
    
    \draw[arrow_style, color=purple!80] (strat_radical.west) -- (act_mem.east);

    \end{tikzpicture}
    } 
    \caption{Decomposition of Memory Constraints. To fit models within limited VRAM, strategies target specific components of the memory equation: Quantization compresses static weights ($M_\theta$), reducing the memory footprint across both training and inference. Conversely, Block-wise Optimization and Data/Radical methods specifically mitigate the dynamic memory overhead of optimizer states ($M_{\mathcal{O}}$) and activation storage ($M_{\mathcal{A}}$) exclusively during the training phase.}
    \label{fig:memory_decomp_onecol}
\end{figure}
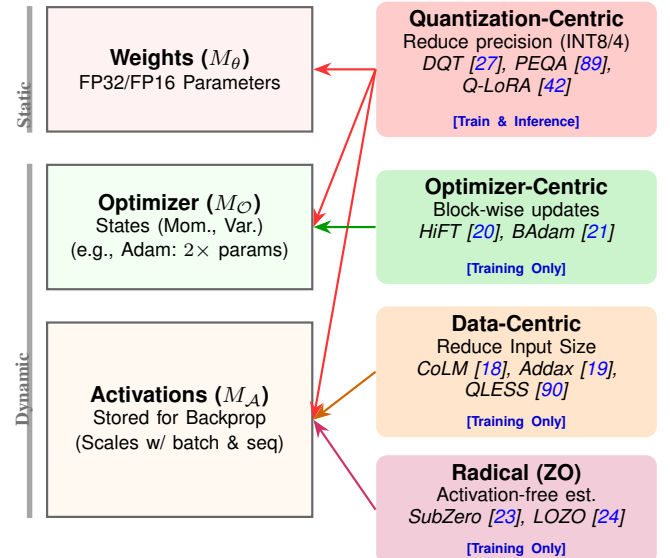

\subsection{The Data-Centric Approach}

A naive approach to reducing activation memory ($M_{\mathcal{A}}$) is to decrease the mini-batch size. However, this leads to heavily noisy gradients and unstable convergence, both theoretically and practically, as the variance of the stochastic gradient scales inversely with the batch size ($1/b$). Recent methods, however, demonstrate that intelligent data selection can circumvent this trade-off.

The CoLM (Coresets for Training LLMs) method addresses the specific problem that large batches are often too memory-intensive for training LLMs, while small batches fail to represent the data distribution adequately \cite{nguyen2025minibatch}. Instead of using random small batches, CoLM iteratively selects small, weighted subsets (coresets) $\mathcal{S}$ with weights $\gamma_z$ to approximate the gradient of a theoretical, much larger batch $b_{large}$:

\begin{equation}
    \sum_{z \in b_{large}} \nabla \mathcal{L}(z; \theta) \approx \sum_{z \in \mathcal{S}} \gamma_z \nabla \mathcal{L}(z; \theta).
\end{equation}

A central aspect here is dealing with data imbalance in language data. Standard selection methods often ignore small data sources. CoLM solves this with a hybrid approach: all examples from small sources are retained, while only representative ``medoids" are selected from large sources. Since LLMs are predominantly trained with Adam~\cite{kingma2014adam}, CoLM does not match raw gradients, but rather the preconditioned gradients $\tilde{g}$. This adapts the selection to the optimizer's variance estimate $\hat{v}_t$:

\begin{equation}
    \tilde{g}(z; \theta_t) = \frac{\nabla \mathcal{L}(z; \theta_t)}{\sqrt{\hat{v}_t} + \epsilon}.
\end{equation}

In practice, CoLM reduces the memory requirement for fine-tuning by a factor of 2 and even outperforms training with randomly selected batches that are four times larger \cite{nguyen2025minibatch}. However, calculating the ``value" of data for selection can itself be memory-intensive. Addressing this, QLESS~\cite{ananta2025qlessquantizedapproachdata} introduces a quantized approach to influence estimation. By using quantized gradient representations to approximate data influence, QLESS substantially reduces the computational and memory overhead of the selection process itself. This expands the regime in which influence-aware selection becomes feasible under fixed memory budgets.

Going a step further is Addax \cite{li2024addax}, a hybrid approach that dynamically optimizes memory requirements based on sequence length. Since the memory required for gradient computation correlates strongly with input length, Addax partitions the dataset based on a length threshold $L_T$. The optimization objective splits into a memory-efficient First-Order (FO) part for short sequences and a memory-intensive Zeroth-Order (ZO) part for long sequences:
\begin{equation}
    \theta_{t+1} = \theta_t - \eta \left( \nabla_{FO}\mathcal{L}(\mathcal{B}_{short}; \theta_t) + \hat{\nabla}_{ZO}\mathcal{L}(\mathcal{B}_{long}; \theta_t) \right).
\end{equation}
For long sequences ($> L_T$), Addax utilizes a Zeroth-Order estimator (MeZO), which estimates the gradient using only two forward passes with a random perturbation $u$, avoiding the storage of activation maps entirely:
\begin{equation}
    \hat{\nabla}_{ZO}\mathcal{L}(z; \theta) \approx \frac{\mathcal{L}(z; \theta + \epsilon u) - \mathcal{L}(z; \theta - \epsilon u)}{2\epsilon} u.
\end{equation}
For short sequences ($\le L_T$), a standard First-Order optimizer (In-Place SGD) is used. This split overcomes the typically slow convergence of pure Zeroth-Order methods. Simultaneously, the Zeroth-Order component acts as a regularizer, helping to avoid sharp local minima. On an A100 setup, Addax was able to successfully fine-tune an OPT-13B model on all tested tasks, whereas standard SGD failed due to ``Out-of-Memory" errors \cite{li2024addax}.

\subsection{The Optimizer-Centric Approach}
If data reduction is not desired, the memory requirement of the optimizer itself must be addressed. Standard algorithms like Adam require additional memory slots for momentum and variance for each parameter (totaling $18M$ memory for a model with $M$ parameters), which is prohibitive for billions of parameters.

Here, HiFT (Hierarchical Full Parameter Fine-Tuning) offers an architectural solution \cite{liu2024hift}. Instead of updating all model parameters simultaneously, HiFT divides the parameters $\theta$ into a set of hierarchies or blocks $\mathcal{B} = \{b_1, \dots, b_k\}$. In each training step, only a subset of parameters $\theta_{b_i}$ (one block) is updated, while the rest remains frozen:
\begin{equation}
    \theta_{b_i}^{(t+1)} \leftarrow \text{Update}(\theta_{b_i}^{(t)}, \nabla_{\theta_{b_i}}\mathcal{L}), \quad \theta_{\setminus b_i}^{(t+1)} \leftarrow \theta_{\setminus b_i}^{(t)}
\end{equation}
This significantly reduces the amount of gradients and optimizer states that must be held in GPU memory simultaneously. Unlike classical layer-wise training, HiFT uses a delayed learning rate update mechanism to ensure end-to-end stability. This method reduces the number of trainable parameters per step by an average of 89.18\% and enables full fine-tuning of a 7B model on a 24GB consumer GPU \cite{liu2024hift}.

The theoretical foundation for such block-wise updates is provided by BAdam \cite{luo2024badam}. It transfers the mathematical principle of Block Coordinate Descent (BCD) to the Adam optimizer. BAdam partitions the model parameters into $D$ disjoint blocks $\mathcal{G}_1, \dots, \mathcal{G}_D$. At each step, it sequentially updates only one active block $\mathcal{G}_k$ to minimize the loss while keeping other blocks fixed:
\begin{equation}
    \min_{\theta_{\mathcal{G}_k}} \mathcal{L}\left(\theta_{\mathcal{G}_k} \cup \theta_{\setminus \mathcal{G}_k}^{(fixed)}\right).
\end{equation}
This is approximated by performing $K$ Adam steps on the active block. Crucial for memory efficiency is that the memory-intensive optimizer states (momentum, variance) are deleted after a block is updated. This reduces the memory requirement $M_{mem}$ drastically compared to standard Adam:
\begin{equation}
    M_{mem}^{BAdam} \approx 2 M + \frac{16 M}{P} \ll 18 M
\end{equation}
Here, $2M$ represents the minimal storage for FP16 weights and gradients, while the optimizer states ($16M$) are divided by the number of blocks $P$. BAdam learns updates with ``full rank", which leads to better performance on complex downstream tasks compared to low-rank methods like LoRA, while maintaining comparable memory requirements \cite{luo2024badam}.

Crucially, this reduction in memory state introduces a fundamental ``time-for-memory" trade-off. Because block-wise methods approximate the global optimization trajectory via BCD, they inherently suffer from update lag; parameters in frozen blocks cannot react immediately to loss changes induced by the active block. Furthermore, the frequent resetting or partitioning of optimizer states disrupts the global momentum history that standard Adam relies on for acceleration. Consequently, while methods like BAdam and HiFT make fine-tuning feasible on consumer hardware, they often require a greater number of training epochs to match the convergence of full-parameter baselines, effectively increasing total wall-clock time to minimize peak memory usage.

Block-wise methods such as HiFT and BAdam reduce the \emph{active} optimizer/gradient state \emph{per step} by updating only a subset of parameters at a time and discarding (or not instantiating) optimizer states for inactive blocks. In contrast, in multi-GPU training, memory is often managed via distributed sharding methods such as ZeRO (Zero Redundancy Optimizer)~\cite{rajbhandari2020zero} and Fully Sharded Data Parallel (FSDP)~\cite{zhao2023pytorch}, which partition parameters/gradients/optimizer states across workers to reduce per-device VRAM. 
As a result, block-wise methods are typically \emph{complementary} to sharding: they are most useful in single-GPU or small-cluster regimes, or when sharding alone is insufficient to fit the active state within per-device VRAM.

\subsection{Gradient-Free Approximation}
The biggest memory consumer for long sequences is not the weights, but the activations cached for backpropagation. Zeroth-Order (ZO) methods eliminate this by estimating gradients via forward pass differences. The standard estimator uses a random perturbation $u \sim \mathcal{N}(0, I)$ and scaling $\epsilon$:
\begin{equation}
    \hat{\nabla}\mathcal{L}(\theta) = \frac{\mathcal{L}(\theta + \epsilon u) - \mathcal{L}(\theta - \epsilon u)}{2\epsilon} u.
\end{equation}
However, the variance of this estimator scales linearly with the number of parameters $N$ ($\text{Var} \propto N$), which leads to instability for LLMs.

The SubZero (Random Subspace Optimization) approach \cite{yu2024subzero} addresses this by optimizing not in the full parameter space $\mathbb{R}^N$, but in a lower-dimensional random subspace $\mathcal{S} \subset \mathbb{R}^N$ defined by a projection matrix $\mathcal{P}$. The gradient is estimated as:
\begin{equation}
    \hat{\nabla}_{\mathcal{S}}\mathcal{L}(\theta) = \frac{\mathcal{L}(\theta + \epsilon \mathcal{P} u) - \mathcal{L}(\theta - \epsilon \mathcal{P} u)}{2\epsilon} \mathcal{P} u
\end{equation}
Technically, SubZero uses a layer-wise perturbation where $\mathcal{P}$ is constructed from small orthogonal matrices. To save compute, $\mathcal{P}$ is "frozen" for $T_0$ iterations (Lazy Update), reducing the effective dimension $N_{eff} \ll N$ and thus the variance \cite{yu2024subzero}.

A further development is LOZO (Low-Rank ZO-SGD) \cite{chen2024enhancing}, which builds on the observation that gradients in fine-tuning possess a low-rank structure. Instead of a full gradient matrix $G$, LOZO assumes $G \approx U V^\top$. The update rule incorporates momentum $m_t$ directly on these factors:
\begin{equation}
    \theta_{t+1} = \theta_t - \eta  (U_t V_t^\top), \quad \text{with } m_t \text{ stored as } (m_U, m_V).
\end{equation}

Since LOZO stores momentum in this compressed form ($m_U, m_V$), the memory overhead for the optimizer states scales linearly with the rank $r$, specifically $O(rN)$. This approach avoids the quadratic complexity of the full model parameters $O(N^2)$ while simultaneously reducing variance compared to random perturbations \cite{chen2024enhancing}.

To further stabilize convergence, AdaZeta \cite{yang2024adazeta} combines ZO methods with Tensor-Train Decomposition. Crucially, to prevent divergence, AdaZeta increases the number of gradient queries $Q$ per epoch sublinearly with the number of iterations $t$:
\begin{equation}
    Q_t \propto t^\nu, \quad \text{with } 0 < \nu < 1.
\end{equation}
This dynamic scaling ensures that the gradient approximation error $||\nabla \mathcal{L} - \hat{\nabla} \mathcal{L}||$ decreases over the course of training, stabilizing the fine-tuning of large models without unnecessarily inflating runtime at the start \cite{yang2024adazeta}.

\subsection{The Quantization-Centric Approach}
\label{sec:quant}

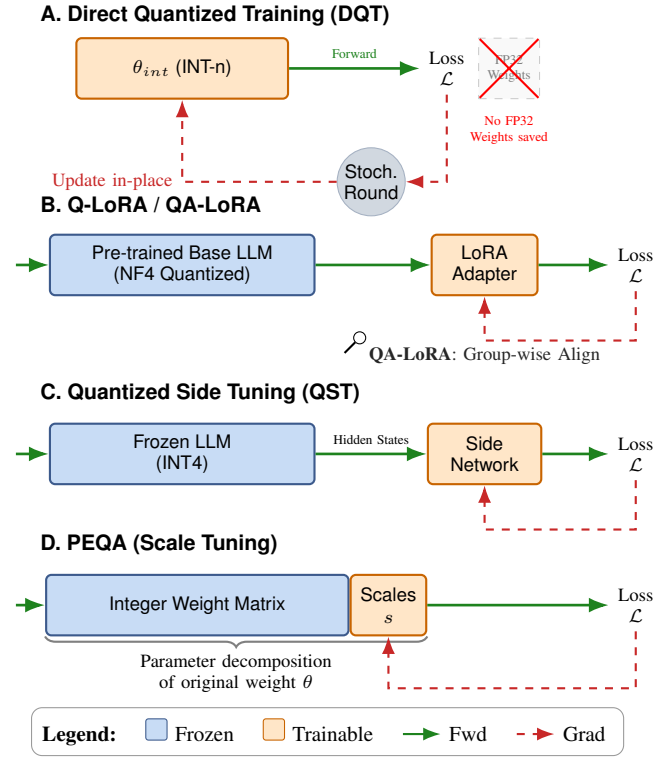
\begin{figure}[htbp]
    \centering
    \input{quantization.tex} 
    \caption{Overview of Memory-Efficient Quantized Learning Methods.}
    \label{fig:quantization}
\end{figure}
While the previous methods address dynamic memory (activations and optimizer states), they assume the model weights $M_{\theta}$ themselves must be stored in high precision (FP32 or FP16).
Direct Quantized Training (DQT)~\cite{zhao2024direct} introduces a memory-efficient training methodology that eliminates the FP32 “master” weights used in conventional Quantization-Aware Training (QAT). Instead of maintaining dual representations of the model FP32 weights for gradient updates and low-bit weights for forward passes, as illustrated in Figure \ref{fig:quantization}A, DQT initializes the model directly in an INT-n format (denoted as $\theta_{\text{int}}$) and keeps it quantized throughout training. During backpropagation, the optimizer produces a temporary high-precision update, but this update is never stored; it is immediately projected back into the low-bit domain using stochastic rounding (SR):
\begin{equation}
    \theta_{\text{int}}^{(t+1)} = \text{SR}\left( \theta_{\text{int}}^{(t)} - \eta \nabla\mathcal{L} \right).
\end{equation}
This probabilistic quantization rule preserves small update signals without requiring differentiable quantization. Typically, for a value $v$, it is defined as rounding to the nearest discrete states based on distance:
\begin{equation}
    \text{SR}(v) = \begin{cases} 
    \lfloor v \rfloor + 1 & \text{with probability } v - \lfloor v \rfloor \\
    \lfloor v \rfloor & \text{with probability } 1 - (v - \lfloor v \rfloor)
    \end{cases}
\end{equation}
This mechanism replaces the Straight-Through Estimator and avoids re-quantizing FP32 weights each step, thereby eliminating the memory overhead associated with both high-precision parameters and their optimizer states. By updating the quantized weights in place, DQT maintains a single low-bit copy of the model at all times, achieving substantial reductions in training-time memory footprint while retaining compatibility with standard backpropagation and optimizers.

Moving beyond full model training, other methods focus on adapting frozen, quantized backbones. QLoRA \cite{dettmers2023qlora} (Figure \ref{fig:quantization}B) reduces memory overhead by backpropagating gradients through a frozen, NF4-quantized base model into trainable Low-Rank Adapters (LoRA), using techniques like Double Quantization to maximize efficiency. However, because QLoRA requires dequantizing weights during computation, it faces challenges with efficient deployment. QA-LoRA \cite{xu2023qaloraquantizationawarelowrankadaptation} addresses this specific limitation by introducing group-wise operators that align the granularity of quantization with the adapter parameters. This structural alignment allows the adapter and base weights to be merged losslessly into INT4, enabling significantly faster inference than QLoRA while maintaining accuracy.

To further reduce the memory cost of intermediate activations, Quantized Side Tuning (QST) \cite{zhang2024quantizedtuningfastmemoryefficient} (Figure \ref{fig:quantization}C) employs a dual-stage framework. Instead of passing gradients through the base model, QST pairs a frozen 4-bit LLM with a lightweight Side Network that uses hidden states from the base model to make predictions. By avoiding backpropagation through the massive base LLM entirely, QST significantly cuts the memory footprint for both activations and optimizer states. 

Finally, Parameter-Efficient and Quantization-aware Adaptation (PEQA) \cite{kim2023memoryefficientfinetuningcompressedlarge} (Figure \ref{fig:quantization}D) bridges the gap between PEFT and quantization through "scale tuning." It decomposes pre-trained weights into a frozen integer matrix and trainable scales ($s$). By updating only these quantization scales, PEQA achieves massive compression—such as reducing a 65B model’s requirement from 131GB to 33GB—while restoring performance competitive with full-precision baselines.

From a fine-tuning efficiency perspective, DQT and post-training adaptation of quantized models optimize different parts of the memory/compute balance. DQT maximally compresses the weight and optimizer footprint during training by keeping a single INT-n copy and never materializing FP32 masters; this removes optimizer-state bloat but injects stochastic quantization noise into every update, which can slow convergence or require higher bitwidths to match accuracy. In contrast, QLoRA, QA-LoRA, and PEQA defer quantization to a frozen backbone and concentrate learning on small adapters or scales. They avoid pervasive quantization noise and are typically easier to tune, but they incur per-step dequantization overhead and higher activation/optimizer costs during backprop, shifting the bottleneck to compute and activation memory. The choice is thus context-dependent: DQT favors the most aggressive VRAM savings when training stability is manageable, whereas adapter/scale-based approaches trade a modest memory increase and compute overhead for more stable optimization.

For edge-compatible LLMs, constraints extend beyond generic memory to strict energy and thermal envelopes, unstable connectivity, and tight tail-latency requirements \cite{rajapakse2023extremeedge,li2024mobilelatency,tan2024thermalaware}. This shifts the objective from maximizing final accuracy alone to balancing accuracy-per-joule, thermal stability, and P95/P99 latency under offline-capable operation \cite{li2024mobilelatency,tan2024thermalaware}. Hence, methods with deployable integer-friendly paths (e.g., QA-LoRA-style mergeable INT4 adapters or PEQA scale tuning) are often preferable for stable on-device inference, while DQT is attractive when on-device adaptation is needed under severe memory limits.

\begin{table*}[t]
\centering
\begin{threeparttable}
\caption{Compact engineering summary of representative efficiency methods\tnote{*}}
\label{tab:compact_tradeoffs}
\setlength{\tabcolsep}{2.2pt}
\renewcommand{\arraystretch}{1.06}
\scriptsize
\begin{tabular}{p{1.3cm} p{1.35cm} p{1.7cm} p{2.15cm} p{2.1cm} p{1.75cm} p{1.7cm} p{1.8cm}}
\toprule
\textbf{Meth.} & \textbf{Target} & \textbf{Mechanism} & \textbf{Memory} & \textbf{Compute} & \textbf{Perf.} & \textbf{Best use} & \textbf{Caveat} \\
\midrule
CoLM~\cite{nguyen2025minibatch} &
$M_{\mathcal{A}}$ &
Coreset mini-batches &
$2\times$ less memory; beats random batches $4\times$ larger &
Selection overhead; cheaper than large-batch training &
Up to +7.1\% / +20\% vs.\ random &
VRAM-limited large-batch FT &
Selection cost grows at scale \\
\midrule
QLESS~\cite{ananta2025qlessquantizedapproachdata} &
Selection cost &
LoRA proj. + low-bit gradient store &
Up to $16\times$ less gradient-store memory &
Extra scoring stage &
Near-LESS quality; 1-bit often sufficient &
Influence-based selection under tight memory &
Improves selector, not optimizer \\
\midrule
Addax~\cite{li2024addax} &
$M_{\mathcal{A}}$ &
FO short + ZO long seqs. &
Up to 89\% memory reduction; enables OPT-13B on 1 A100 &
$15\times$ / $30\times$ faster than MeZO on reported setups &
+14\% and $>$16\% over MeZO &
Long-context FT &
ZO part can be noisy \\
\midrule
HiFT~\cite{liu2024hift} &
$M_{\mathcal{O}}$ &
Hierarchical or blockwise update &
89.18\% fewer trainable params/step; 7B on 24\,GB GPU &
More complex update schedule &
Comparable to PEFT/full FT &
Commodity-hardware full FT &
Tuning/scheduling complexity \\
\midrule
BAdam~\cite{luo2024badam} &
$M_{\mathcal{O}}$ &
Block-coordinate Adam &
$18M \rightarrow 2M+16M/P$ &
Extra block scheduling; efficient backward &
On par with/better than Adam, better than LoRA &
Low-memory full FT &
Sensitive to block partition \\
\midrule
QLoRA~\cite{dettmers2023qlora} &
$M_\theta$ &
4-bit frozen base + LoRA &
65B FT on single 48\,GB GPU &
Dequantization overhead &
Near 16-bit FT performance &
Default VRAM-limited PEFT &
Less deployment-efficient \\
\midrule
QA-LoRA~\cite{xu2023qaloraquantizationawarelowrankadaptation} &
$M_\theta$ &
Quant.-aware LoRA, mergeable INT4 &
Low-bit FT with INT4 mergeability &
Faster deployment than QLoRA &
Accuracy retained with better deployability &
On-device / low-bit deployment &
More specialized design \\
\midrule
PEQA~\cite{kim2023memoryefficientfinetuningcompressedlarge} &
$M_\theta$ &
Frozen integer weights + scale tuning &
131\,GB $\rightarrow$ 33\,GB; 4--5$\times$ smaller &
Low update cost &
Competitive up to 65B &
Deployment-oriented tuning &
Limited adaptation capacity \\
\midrule
DQT~\cite{zhao2024direct} &
$M_\theta,M_{\mathcal{O}}$ &
In-place quantized updates; no FP32 master copy &
Removes FP32 master + optimizer overhead &
Less requantization; noisier optimization &
About 5\% loss degradation at 8-bit vs.\ cited baseline &
Aggressive training-time VRAM reduction &
Harder optimization stability \\
\bottomrule
\end{tabular}
\begin{tablenotes}[flushleft]
\footnotesize
\item[*]Reported effects are taken from the original papers and are not directly comparable across models, datasets, hardware, or training setups.
\end{tablenotes}
\end{threeparttable}
\end{table*}

\subsection{Synthesis: Towards a Unified View of Memory-Efficient Fine-Tuning}

The progression of methods surveyed in this section (summarized in Table~\ref{tab:compact_tradeoffs}) highlights a fundamental shift in LLM training: the move from hardware-centric scaling to algorithmic efficiency. However, a critical analysis reveals that current solutions generally operate in isolation, targeting only one term of the memory equation ($M_{\theta}$, $M_{\mathcal{O}}$, or $M_{\mathcal{A}}$) while leaving the others as dominant bottlenecks.

Each strategy achieves memory reduction by sacrificing a specific property of standard optimization. Data-centric approaches trade \textit{sample density} for activation space, accepting higher variance to fit batch constraints. Optimizer-centric methods trade \textit{update synchronicity} for state reduction, serializing computations that were previously parallel. Quantization approaches trade \textit{numerical precision} for static storage, risking representational capacity for footprint.

Crucially, these trade-offs are interconnected. For instance, aggressively quantizing weights ($M_{\theta}$) effectively clears static memory, but this often exposes the activation stack ($M_{\mathcal{A}}$) as the new, prohibitive limit for long-context reasoning. Similarly, while block-wise optimizers eliminate state overhead ($M_{\mathcal{O}}$), they do not inherently resolve the activation costs associated with large batches. Consequently, relying on an isolated approach often yields diminishing returns; solving one constraint merely shifts the failure point to another component of the memory equation.

This analysis suggests that the future of efficient fine-tuning does not lie in optimizing a single lever but in the simultaneous compression of all three memory terms. As proposed in recent discussions, the most promising research direction is the structural hybridization of these mechanisms. A unified pipeline combining blockwise optimization to minimize $M_{\mathcal{O}}$ with quantized weights to minimize $M_{\theta}$ and zeroth-order regularization to limit $M_{\mathcal{A}}$ could theoretically distribute the compression load across all variables. Such a holistic approach would effectively decouple model scale from hardware limitations. This enables the fine-tuning of models exceeding 70B parameters on consumer hardware, where no single method could succeed alone.

Hypothesizing a unified pipeline stacking DQT's SR with Addax's zeroth-order (ZO) estimators for long sequences reveals key noise dynamics. SR introduces stochastic quantization noise during in-place INT-n weight updates, which perturbs forward-pass losses and compounds ZO's inherently high-variance finite-difference approximations~\cite{shang2025fine,zhao2024direct}.
This compounding is characteristic of quantized training coupled with ZO methods, where discrete weight errors inflate estimator instability~\cite{shang2025fine}. Recent work confirms the potential for such noise accumulation but demonstrates effective mitigations: Shang et al.~\cite{shang2025fine} (QZO) fine-tune quantized networks by perturbing continuous quantization scales with directional clipping, reporting 18$\times$ total memory savings on Llama-2-13B~\cite{touvron2023llama} across tasks, while Bar \& Giryes~\cite{bar2025zoqo} (ZOQO) enable fully quantized ZO-SignSGD via discrete noise injection and scaled LR, achieving $\sim$90\% accuracy on OPT-1.3B-LoRA (SST2) at 8-bit with 60\% memory reduction vs.\ QAT.
Complementary techniques like sparse perturbations (Sparse-MeZO) target noise-resilient parameter subspaces, further enabling hybridization with block-wise optimizers (HiFT/BAdam)~\cite{liu2024sparse, zhou2025quzo}. These advances support simultaneous compression of all memory terms ($M_\theta$, $M_O$, $M_A$), informing the compute governor policy $\pi(S_t, B_t)$ for resource-constrained fine-tuning~\cite{shang2025fine, bar2025zoqo}.


\section{The Governor: Compute Budget Awareness (The ``When" to Stop)}\label{compute}
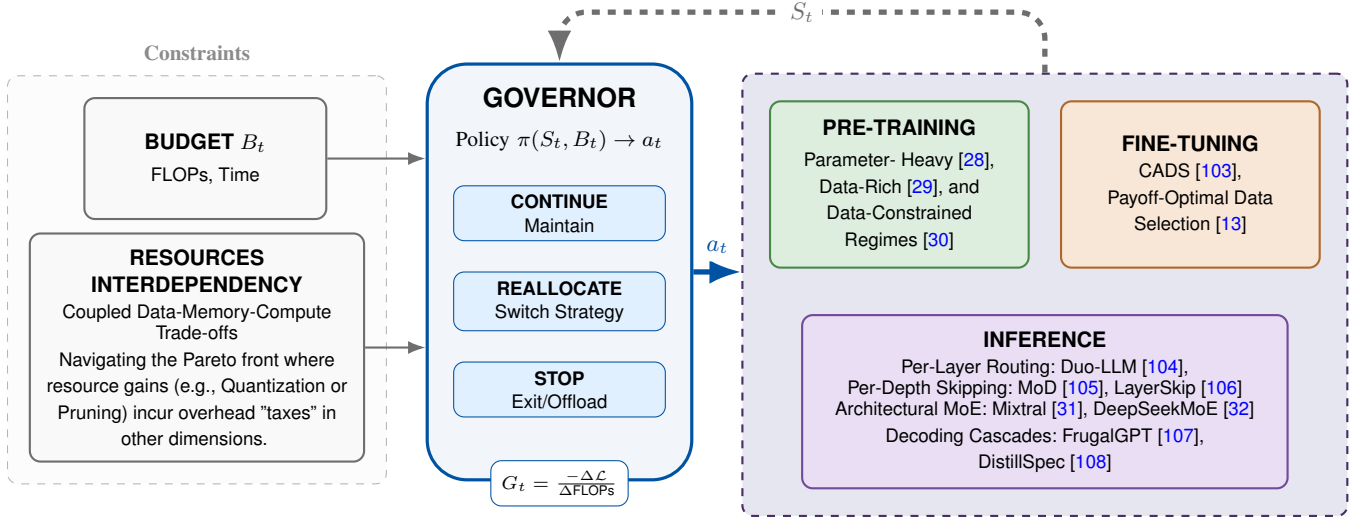
\begin{figure*}[htbp]
    \centering
    \input{governor.tex} 
    \caption{Compute governor coordinating budget-aware decisions across training, fine-tuning data selection, and inference given the resources interdependency. }
    \label{fig:compute-governor}
\end{figure*}

To unify the disparate efficiency techniques discussed thus far, we introduce the concept of a ``compute governor" (or simply ``governor"), an integrated decision-making layer that manages the coupling of data, memory, and compute, rather than treating these bottlenecks as independent variables across training and inference.
We formalize the governor as a control policy:
\begin{equation}
    \pi(S_t, B_t) \to a_t,
    \label{eq:policy}
\end{equation}
that maps the current system state $S_t$ including loss dynamics/gradient statistics and the currently active data and memory strategies together with the remaining resource budget $B_t$ (FLOPs, energy, wall-time, data tokens, or memory capacity) to an action $a_t$. We treat the marginal gain per FLOP, $G_t$, as a feedback signal for the governor:
\begin{equation}
G_t = -\frac{\Delta \mathcal{L}_t}{\Delta \mathrm{FLOPs}},
\end{equation}
i.e., the expected reduction in validation (or training) loss per unit of additional compute.  The governor selects \(a_t \in \{\text{continue}, \text{reallocate}, \text{stop}\}\), where this action determines whether the current configuration is maintained, updated, or terminated as the system evolves to the next state. The governor stops (or reallocates) when \(G_t\) falls below a budget-dependent threshold. An illustration of the governor is provided in Figure~\ref{fig:compute-governor}.

Subsection~\ref{sec:governor-example} introduces a case study of this framework. Subsection~\ref{data-compute} focuses on compute budgeting in fine-tuning via compute-aware and payoff-optimal data selection, \ref{Memory-Compute} trade-offs on the memory--compute axis (e.g., quantization vs.\ dequantization overhead), \ref{scalinglaws} shifts to macro-level compute allocation in pre-training by synthesizing compute-optimal scaling laws across regimes (allocating a FLOPs budget across model size, token budget, and training horizon), and \ref{inference} treats inference as a budgeted per-token computation problem (routing/skipping/MoE/decoding). Subsection~\ref{compute-synthesis} concludes this section.

\subsection{Case Study: An Instantiation of the Compute Governor}
\label{sec:governor-example}
To make the governor in Eq.~(\ref{eq:policy}) more concrete, we illustrate it through a simple budget-aware fine-tuning scenario that jointly coordinates \emph{data selection} and \emph{memory-efficient optimization}. The goal of this example is not to introduce a new training algorithm, but to show how the governor can operate as a practical decision layer over techniques already discussed in Sections~\ref{data} and~\ref{memory}.

At step $t$, the governor observes a compact system state
\[
S_t = (\tilde{G}_t,\; \sigma_t),
\]
where $\tilde{G}_t$ is the current estimate of marginal gain per FLOP, and $\sigma_t=(\sigma_t^{\mathrm{data}},\sigma_t^{\mathrm{mem}})$ represents the current data and memory strategies, respectively. Based on this state, the governor selects an action
\[
a_t \in \{\text{continue},\; \text{reallocate},\; \text{stop}\}.
\]

While $\tilde{G}_t$ measures the marginal gain per FLOP under the current configuration, the governor’s decision requires comparing alternative actions when reallocation is triggered. This is achieved by estimating action-conditioned gains $\tilde{G}_t(a)$, defined as the expected loss reduction per unit compute if action $a$ is applied (e.g., data refresh or block switch). When $\tilde{G}_t$ falls below a budget-dependent threshold, the governor evaluates candidate actions and selects the one with the highest estimated marginal gain, or terminates if no candidate yields sufficient improvement under the remaining budget $B_t$.

In this case study, the \emph{continue} action means training on the current selected subset using the current active parameter block; the \emph{reallocate} action means changing either the data strategy or the memory strategy; and the \emph{stop} action means terminating training once additional compute is no longer justified by the expected loss reduction. This directly instantiates the budget-aware stopping principle described in Section~\ref{compute}, where computation is halted or redirected once its marginal contribution falls below a budget-conditioned threshold.

For the data component, the governor can use an online utility estimate inspired by GREATS~\cite{wang2024greats} to determine whether the current subset remains worthwhile or whether data scores should be refreshed. GREATS is suitable here because it explicitly approximates the marginal validation benefit of candidate examples during training, making it a natural mechanism for estimating whether a new subset is likely to improve utility under a limited compute budget. For the memory component, the governor can use a block-wise optimizer such as BAdam~\cite{luo2024badam}, which reduces optimizer-state memory by updating only one parameter block at a time. These two components operate under the same governor signal: the first controls \emph{which data are worth processing}, while the second controls \emph{which parameters can be updated within memory limits}.

A practical policy can then be stated in simple terms:
\begin{itemize}
    \item If $\tilde{G}_t$ and $B_t$ remain high, the governor chooses \textbf{continue}.
    \item If $\tilde{G}_t$ falls below a budget-dependent threshold, the governor enters a reallocation phase in which it evaluates candidate actions (e.g., refreshing the data subset or switching the active parameter block) by estimating their action-conditioned gains $\tilde{G}_t(a)$. 
    \item If $\tilde{G}_t$ decreases because the current subset has become less informative, and yields lower utility than alternative data selections, but memory remains available, the governor chooses \textbf{reallocate} by refreshing or changing the active data subset.
    \item If memory becomes the binding constraint, and alternative parameter updates offer higher utility, the governor chooses \textbf{reallocate} by switching to a different parameter block or a cheaper memory-saving strategy, while keeping training active.
    \item If no feasible reallocation yields sufficient utility under the remaining budget, the governor chooses \textbf{stop}.
\end{itemize}

This example clarifies the role of the governor in the unified framework. It does not replace the underlying selection or optimization methods; rather, it coordinates them using a shared feedback signal, $\tilde{G}_t$, together with the remaining compute budget and the current hardware state. In this sense, the governor transforms the high-level objective of “maximize performance under constrained resources” into a concrete sequence of decisions about whether to keep training, switch strategy, or terminate.

More broadly, the same pattern extends beyond this example. Different data selectors, memory-reduction methods, or quantization strategies can be substituted into the same decision loop, provided that they expose their expected utility and resource cost to the governor. This is why we view the compute governor as a unifying systems abstraction: it links the dynamic data valuation discussed in Section~\ref{data} with the memory-feasibility mechanisms of Section~\ref{memory} under the common objective of maximizing marginal utility per unit of compute.

\subsection{Data-Compute Pillar: Compute-Aware \& Payoff-Optimal Data Selection} \label{data-compute}
While traditional data selection methods assume that optimal subsets depend solely on informativeness, recent work highlights computational budget as a missing but essential variable. Even when data is perfectly curated (Section~\ref{data}) and fits within memory constraints (Section~\ref{memory}), real-world training is almost always bounded by time, hardware, or financial cost. Thus, the practical objective is not simply to achieve the best possible accuracy, but rather the best accuracy achievable under a fixed compute budget. 

The CADS framework~\cite{wan2025computational} formalizes this idea by treating data selection as a dynamic bilevel optimization problem, where the chosen data must co-evolve with the amount of compute available: In the inner loop, the model is trained on a selected subset of training data subject to a given computational budget.
In the outer loop, data selection is optimized based on the trained model's evaluation. Their findings show that the “optimal’’ data subset is not fixed: under a limited compute budget, models benefit more from easier data, rich in low-frequency features, whereas with larger budgets, harder or more diverse samples (high-frequency information) become advantageous. This compute-dependent trade-off demonstrates that data selection should be explicitly budget-aware, reinforcing that compute constraints fundamentally shape which data is most valuable for training. The formalization of the problem is as follows: 

\begin{equation}
  \begin{aligned}
    \min_{\mathcal{S}} \quad
      & \mathcal{L}_{\mathrm{val}}\bigl(\theta_C(\mathcal{S})\bigr) \\[0.4em]
    \text{s.t.} \quad
      & \theta_C(\mathcal{S})= \text{Train}\bigl( \mathcal{S}, C\bigr), 
  \end{aligned}
  \label{eq:compute_constrained_opt}
\end{equation}
where $C$ is the compute limit, $\mathcal{L}_\mathrm{val}$ is the validation loss, and $\mathcal{S}\subseteq\mathcal{D}$ is a subset of the training set.  
The constraint $\theta_C(\mathcal{S})=\text{Train}(\mathcal{S}, C)$ represents the model parameters derived from training with budget $C$ using the selected subset $\mathcal{S}$. 
Thus, $C$ controls the training horizon (compute), rather than assuming convergence of the inner problem.
Optimizing the discrete subset $\mathcal{S}$ directly is non-differentiable and computationally expensive.
To address this, the authors propose a learnable sampling distribution, parameterized by $s$, from which a binary selection mask $m$ (representing $\mathcal{S}$) is sampled. 
Instead of differentiating through the training trajectory or relying on implicit differentiation, which requires convergence to a local minimum, the method optimizes the distribution parameters $s$ using policy gradients. 
This approach maximizes the expected performance over the distribution of subsets, allowing the model to navigate the bilevel optimization landscape without computing intractable gradients for the inner loop.

Recent work by Yin et al.~\cite{yincompute} takes this one step further and formalizes the trade-off between the computational cost of data selection and the efficiency gains during training. They introduce a framework that explicitly accounts for the total compute budget, encompassing both the overhead of selecting data ($C_v$) and the cost of subsequent model training ($C_T$):  \begin{equation}
\begin{aligned}
\min_{\mathcal{S}} \; & \mathcal{L}_{val}\!\left(\theta(\mathcal S)\right) \\
\text{s.t.} \quad & C_{T(\mathcal S)} + \sum_{x \in \mathcal{D}} C_{v(x)} \le C,
\end{aligned}
\label{eq:compute_constrained_selection}
\end{equation}
where $\mathcal S \subseteq \mathcal{D}$ is a selected subset of the full training dataset $\mathcal{D}$, and $\theta(\mathcal S)$ represents the model parameters trained on $\mathcal S$. $C_{T(\mathcal S)}$ corresponds to the computational cost of training on subset $\mathcal S$, $C_{v(x)}$ denotes the computational cost of computing the utility function $v(x)$ for a sample $x \in \mathcal{D}$, and $C$ represents the total computational budget (e.g., measured in FLOPs) allocated for both data selection and model training.
Their analysis identifies a “pay-back” threshold: perplexity- and gradient-based data selection become efficient only when the training-to-selection model size ratio is approximately 5× and 10×, respectively. This yields a concrete heuristic: high-cost selection methods are justified only when their cost can be amortized by sufficiently large downstream training. For example, they should be used in settings
with repeated training with different tasks on the same underlying models. In LLM settings, this amortization can also be achieved by performing selection with smaller models to enable more efficient training of larger models from the same family; otherwise, under limited budgets, simple heuristics or even random sampling remain more compute-efficient. RHO-1~\cite{lin2025rho1tokensneed} provides complementary evidence at the token level, demonstrating that inexpensive filtering can improve pretraining compute efficiency, aligning with the conclusion that low-cost selection dominates when amortization is limited.


Together, these works demonstrate that data selection for LLMs is inseparable from compute budgeting: both the value of data and the cost of identifying it determine which selection strategies are optimal in practice.

\subsection{Memory-Compute Pillar: Quantization as a Budget-Aware Trade-off}\label{Memory-Compute}

Quantization significantly reduces memory use but does not inherently guarantee compute or latency benefit~\cite{zhao2024atom}. In practice, low‑bit quantization often requires on‑the‑fly dequantization or managing scaling factors, which can add overhead and reduce throughput compared to full‑precision baselines when using generic kernels~\cite{licardo2025performance, park2022lut}. However, with hardware‑optimized kernels that fuse dequantization into low‑bit operations or use native low‑precision arithmetic, this overhead can be largely eliminated and practical speedups realized~\cite{ashkboos2024quik}. As a result, the effect of quantization on marginal gain per FLOP remains dependent on whether memory or compute/latency is the binding constraint in a given deployment~\cite{park2022lut}.

In this context, QLESS~\cite{ananta2025qlessquantizedapproachdata} demonstrates the data-memory-compute trilemma interaction: by quantizing the LESS gradient datastore to 1--8 bits (16$\times$ memory reduction), it enables data valuation under VRAM constraints while preserving selection quality, as confirmed by QLoRA ablation studies. This memory-compressed data selection maintains marginal FLOP utility, providing the governor with a concrete lever: when memory is the binding constraint, prioritize quantized selection over full-precision training to optimize system-level compute efficiency.

\subsection{Compute-Optimal Scaling Laws for Training: Parameter-Heavy, Data-Rich, and Data-Constrained Regimes}\label{scalinglaws}

Kaplan et al.~\cite{kaplan2020scaling} provide the first large-scale empirical scaling laws for language models, showing that performance, as measured by pre-training loss \(L\), improves approximately as a power law in model size \(N\) (number of parameters), dataset size \(D\) (tokens), and total training compute \(C\): 
\begin{equation}
L(N) = \left(\frac{N_c}{N}\right)^{\alpha_N},   
L(D) = \left(\frac{D_c}{D}\right)^{\alpha_D},  
L(C) = \left(\frac{C_c}{C}\right)^{\alpha_C}.    
\end{equation}

Under their fitted exponents $\alpha_N \approx 0.076$, $\alpha_D \approx 0.095$, and $\alpha_C \approx 0.050$, compute-optimal training favors very large models trained on relatively fewer tokens with early stopping, making the recommended regime strongly parameter-heavy. It motivated the trend of extremely large models like GPT-3. 
A regime also exemplified by large models such as Gopher~\cite{rae2021scaling}. This picture is later revised by works such as Hoffmann et al.~\cite{hoffmann2022training}, and Muennighoff et al.~\cite{muennighoff2023scaling}, which argue for a more balanced, data-rich notion of compute-optimality.
The precise numerical values of $N_c$, $D_c$ and $C_c$
depend on the
vocabulary size and tokenization, and hence do not have a fundamental meaning. They emphasize that no plateau was observed within the tested ranges, hinting at continued gains from scaling.
Building on Kaplan et al., Hoffmann et al.~\cite{hoffmann2022training} refine the notion of compute‑optimal training by revisiting the following problem:
\begin{equation}
\begin{aligned}
\min_{N, D} \quad & L(N, D) \\
\text{s.t.} \quad & \text{FLOPs}(N, D) = C.
\end{aligned}    
\end{equation}
By analyzing scaling laws across hundreds of runs, they show that, for a fixed FLOP budget, model size \(N\) and the number of unique training tokens \(D\) should increase roughly in proportion, revealing that many existing large language models are over‑sized and under‑trained relative to their compute. The authors validate this prediction empirically by training Chinchilla, a 70B‑parameter model on 1.4 trillion tokens, which outperforms much larger models trained on fewer tokens while using the same compute budget. This demonstrates that “how far to train” is not arbitrary but follows a concrete scaling law, establishing a reference frontier for compute-optimal training in the abundant-data regime. Subsequent work in data-constrained or multi‑epoch settings was done by Muennighoff et al.~\cite{muennighoff2023scaling}. 
Muennighoff et al.~\cite{muennighoff2023scaling} start from a Chinchilla-style parametric fit,
\begin{equation}
L(N,D) = \frac{A}{N^{\alpha}} + \frac{B}{D^{\beta}} + E,
\end{equation}
and then replace $N$ and $D$ by effective quantities that discount repeated data in the data-constrained regime:
\begin{equation}
L(N,D) = \frac{A}{(N')^{\alpha}} + \frac{B}{(D')^{\beta}} + E,
\end{equation}
where $D'$ is defined via an exponential decay with repetition (see Eq.~(5) in~\cite{muennighoff2023scaling}) and $N'$ analogously (see Eq.~(6) in~\cite{muennighoff2023scaling}).
Building on this baseline, Muennighoff et al.~\cite{muennighoff2023scaling} extend compute‑optimal scaling to explicitly data‑constrained regimes, where the available pool of unique tokens is fixed and additional compute must be spent on repeating data and/or enlarging the model. They note that the trend of increasing both model parameters and dataset size may soon be limited by the finite availability of text data on the internet. The revisited problem is as follows:
\begin{equation}
\begin{aligned}
\min_{N, D} \quad & L(N, D), \\
\text{s.t.} \quad & \text{FLOPs}(N, D) = C\\
& U_D \leq D_C,
\end{aligned}    
\end{equation}
in which $U_D$ is the number of unique tokens used, and $D_C$ is the data budget.
They introduce a modified scaling law that replaces raw tokens and parameters with “effective” quantities that decay as data is repeated and as model size overshoots what the data can support, formalizing the intuition that both repeated tokens and excess parameters exhibit sharply diminishing returns. Empirically, under fixed data pools and specific architectural and optimization regimes, these results indicate that allocating additional FLOPs to increased training duration on smaller models can be compute-competitive with scaling model size or introducing limited amounts of new data. Importantly, this behavior is regime-dependent and degrades as data diversity decreases or model capacity becomes misaligned with the task.

Regime definitions and practitioner guidance: Following Chinchilla~\cite{hoffmann2022training}, we define the \emph{data-abundant} regime as settings achieving $\approx 20$ tokens per parameter (TPP) via single-pass training on unique data, or equivalently $\sim 5$ TPP with up to 4 epochs repetition (negligible 0.5\% loss penalty~\cite{muennighoff2023scaling}). The \emph{data-constrained} regime requires $\gg 4$ epochs on limited unique corpora, where Muennighoff et al.~\cite{muennighoff2023scaling} show repetition value decays rapidly via a modified scaling law $L(N, D|D_C) \propto 1/N'^\alpha + 1/D'^\beta$. Kaplan's parameter-heavy scaling~\cite{kaplan2020scaling} applies to idealized abundant-data, early-stopping cases; Chinchilla corrects to balanced $N \propto D$; data-constrained favors more epochs over parameters until plateau. \emph{Practitioner recipe:} abundant-data → Chinchilla proportionality; constrained → prioritize epochs; avoid Kaplan universally.



\subsection{Budget-Aware Inference: KV Cache Efficiency and Hierarchical Compute Allocation}
\label{inference}

Budget-aware inference spans both memory-system mechanisms (e.g., KV-cache management/compression) and compute-allocation mechanisms (routing/depth/expert/decoding decisions).


\subsubsection{KV Cache Efficiency (memory budget)}
As batch sizes and context lengths grow during inference, the memory bottleneck shifts from model weights to the
key-value cache (KV cache), which stores KV states for all prior tokens across layers; in large batch/sequence regimes the
paper shows KV cache loading can dominate weight loading (e.g., batch 512, context 2048) and even reach $\sim$3$\times$
parameter size in a 500B-class example~\cite{pope2023efficiently, liu2024kivi}. Accordingly, inference efficiency is regime-dependent (prefill vs. decode) and often memory/communication-bound rather than compute-bound.

Techniques like PagedAttention (vLLM)~\cite{kwon2023efficient} manage the KV cache with OS-inspired paging: they store KV states in fixed-size blocks that can be non-contiguous, which reduces internal/external fragmentation and enables near-zero KV cache waste (and block-level sharing), translating into about 2–4× higher serving throughput at similar latency in their evaluation. They also quantify that in existing systems, only about 20.4\%–38.2\% of KV cache memory stores actual token states (the rest is reservation/fragmentation/other waste), while vLLM reaches about 96.3\% token-state usage in the shown experiment.

KIVI~\cite{liu2024kivi} proposes a tuning-free asymmetric 2-bit KV-cache quantization scheme that quantizes the key cache per-channel and the value cache per-token.
With a hardware-friendly implementation, they report 2.6× lower peak memory (including model weights) while maintaining almost the same quality for Llama/Falcon/Mistral.
They further report that this memory reduction enables up to 4× larger batch size, yielding 2.35–3.47× throughput improvement on a real LLM inference workload.


\subsubsection{Hierarchical Compute Allocation (compute/latency budget)}
The Duo-LLM framework~\cite{alizadeh2024duo} augments each feed-forward layer with a small auxiliary module alongside the original large FFN and studies, via an oracle upper bound, how to route tokens between small, large, or skipped modules under a fixed FLOP budget per token. Using this oracle, the authors show that budget-optimal routing patterns are highly non-trivial: for example, activating a large module in only a small subset of layers can yield lower perplexity than using large modules in all layers, and conventional learned routers substantially underutilize the available compute compared to these oracle optima~\cite{alizadeh2024duo}. This perspective extends compute-optimality from global choices over model size, data, and training duration to fine-grained, per-token allocation of compute within the network at inference time, underscoring that ``when and where to spend compute'' must be made explicitly budget-aware across both training and decoding. 

While Duo-LLM focuses on per-token routing within layers, learned routing strategies can approximate the oracle by training a router to assign tokens to modules based on input complexity. Sparse MoE architectures such as Mixtrall~\cite{jiang2024mixtralexperts} and DeepSeekMoE~\cite{dai2024deepseekmoeultimateexpertspecialization} implement conditional computation at the architectural level, though they typically use fixed routing policies rather than optimizing per-token compute budgets.

Furthermore, recent advances in dynamic routing challenge the assumption that every token requires the full depth of the model. Approaches such as \textit{Mixture-of-Depths} \cite{raposo2024mixtureofdepthsdynamicallyallocatingcompute} and \textit{LayerSkip} \cite{Elhoushi_2024} enforce explicit or expected per-token compute budget by allowing ``easy" tokens to bypass layers or exit early, thereby allocating the majority of FLOPs only to the most complex segments of the generation process.

Speculative decoding fundamentally accelerates inference by mitigating the bottleneck of autoregressive generation, where the massive target model typically executes fully for every single token. Instead, this paradigm employs a lightweight ``draft" model to rapidly hypothesize a sequence of tokens (e.g., 5 steps) at negligible cost, allowing the target model to verify all candidate tokens in a single parallel forward pass; if the draft matches the target, multiple tokens are generated for the cost of one \cite{leviathan2023fast, chen2023accelerating}. Recent frameworks extend this principle to maximize budget efficiency: \textit{FrugalGPT} \cite{chen2023frugalgpt} implements a ``cascade" strategy that first queries cheaper, lower-capacity models (e.g., GPT-3.5) and escalates to expensive models (e.g., GPT-4) only for low-confidence queries, while \textit{DistillSpec} \cite{zhou2023distillspec} enhances the acceptance rate of the draft model by specifically distilling the target model's behavior into the drafter, ensuring higher alignment and faster effective generation.

As with training-time allocation, these inference-time strategies define regime-dependent optima shaped by latency constraints, model architecture, and deployment objectives rather than universal prescriptions. Training-time compute-efficient choices constrain which inference-time budget strategies are available and how well they work.
Architectural decisions such as adding experts/auxiliary modules and learning a router determine whether conditional computation (per-layer routing or MoE) can be applied at inference at all, and the routing objective used during training can bias the specialization and effectiveness of cheaper modules, affecting the attainable budget--quality trade-off at inference~\cite{alizadeh2024duo}.
Likewise, decoding-time acceleration via speculative decoding depends on training: the achievable speedup is governed by the draft--target alignment (and thus the acceptance rate), which can be improved by distilling the target into the drafter (e.g., DistillSpec), directly coupling training choices to inference-time budget efficiency~\cite{zhou2023distillspec}.
Finally, compute-optimal training choices over model size versus data (e.g., Chinchilla-style scaling) set the baseline inference cost envelope (FLOPs and memory footprint), which in turn influences whether deployments are compute-limited or KV cache/memory-traffic-limited in a given batch/context regime~\cite{hoffmann2022training}.

\subsection{Synthesis: When and Where to Spend Compute}
\label{compute-synthesis}

The coupled effects of data-, memory-, and compute-efficiency mechanisms are summarized in Table~\ref{tab:coupling_analysis}. This table provides a compact summary of the cross-bottleneck trade-offs that motivate the compute governor.
Across the compute-aware data selection approaches, stopping criteria are naturally defined in terms of compute efficiency rather than convergence. Training or selection is typically terminated when the marginal performance gain per unit of additional compute falls below a budget-dependent threshold, or when the remaining compute is insufficient to amortize the cost of further data valuation or filtering. In this view, stopping is not universal but regime-specific: under tight budgets, early stopping or coarse selection is optimal, whereas larger budgets justify prolonged training and more expensive, fine-grained selection. This perspective further reinforces that data selection and stopping rules are inseparable from explicit compute constraints.

Moreover, the scaling laws induce a practical stopping criterion: training should cease once additional compute would push the system beyond the compute-optimal frontier, where either repeated data or excess parameters yield sharply diminishing returns in loss per FLOP.

Finally, at inference time, the governor manifests as a per-token stopping rule that decides when additional layers, experts, or verification steps are no longer justified by expected uncertainty reduction. Inference-time stopping is typically governed by confidence thresholds, acceptance criteria, or fixed latency budgets, further reinforcing the view of decoding as a budget-constrained decision process.
For concreteness at inference, speculative decoding accepts draft tokens via verification matching~\cite{chen2023accelerating}, models use entropy-based confidence thresholds~\cite{kadavath2022language}, and latency budgets employ adaptive KV eviction (TimeBill~\cite{fan2025timebill}).

Across data selection, training, and inference, a common principle emerges: computation should be halted, or reallocated, once its marginal contribution to performance falls below a budget-conditioned threshold. In this sense, stopping is not a fixed epoch count or convergence criterion, but a dynamic decision governed by diminishing returns under computational constraints.

\begin{table*}[t]
\centering
\begin{threeparttable}
\caption{Cross-bottleneck impact analysis: how pillar-level efficiency levers couple \emph{data}, \emph{memory}, and \emph{compute}. In the Data column, $\uparrow$ denotes higher learning utility per token; in Memory/Compute columns, $\downarrow$ denotes reduced resource use.}
\label{tab:coupling_analysis}
\renewcommand{\arraystretch}{1.2}

\begin{tabular}{l p{4cm} p{4cm} p{4cm}}
\toprule
\textbf{Technique} & \textbf{Data Efficiency} & \textbf{Memory Footprint} & \textbf{Compute / Convergence Impact} \\
\midrule
Data selection & High \textcolor{Green}{$\uparrow$} & Mixed (data selection \textcolor{red}{$\uparrow$}; training \textcolor{Green}{$\downarrow$}) & Mixed (data selection \textcolor{red}{$\uparrow$};  training \textcolor{Green}{$\downarrow$})\\
Memory efficiency & Indirect\tnote{*} \textcolor{Green}{$\uparrow$} & High \textcolor{Green}{$\downarrow$} & Hardware-dependent\tnote{**} \\
Compute efficiency & Regime-dependent\tnote{a} & Mixed\tnote{b} & High \textcolor{Green}{$\downarrow$} \\
\bottomrule
\end{tabular}

\begin{tablenotes}[flushleft]
\footnotesize
\item[*] might enable longer context/batch; more diverse data mixtures

\item[**] often \textcolor{red}{$\uparrow$} FLOPs/wall-time (recompute, update lag, extra forward passes); can be approximately neutral with HW/SW co-design
\item[a] tight budgets favor cheaper/easier data; expensive selection only if amortized
\item[b] may shift bottleneck to VRAM/KV cache; may require cache management for throughput 
\end{tablenotes}
\end{threeparttable}
\end{table*}

\section{Conclusion}\label{conclusion}
This survey argued that making LLMs truly ``efficient'' requires treating data, memory, and compute as a single, coupled optimization problem rather than three independent axes. Building on recent advances in data-centric training, memory-efficient optimization, and compute-aware governance, it showed how each lever, data selection, memory reduction, and budgeted computation, redefines what is achievable under fixed hardware and cost constraints, but also how optimizing only one dimension merely shifts the bottleneck to the remaining terms of the training and inference pipeline.
To move towards truly autonomous edge intelligence, future research must transition from hardware-agnostic efficiency to hardware-aware co-design, where data selection and memory compression are dynamically tuned to the specific energy and latency envelopes of the target edge platform.

The data-centric literature reveals a shift from static pruning and noise filtering (e.g., LIMA, AlpaGasus) toward dynamic, influence-aware selection that maximizes marginal utility per token, yet also exposes a “selection paradox”: sophisticated influence-based methods offer strong theoretical gains but can be prohibitively expensive to run at the scales where LLMs operate, especially when recomputing scores over evolving model states. Methods such as LESS, GREATS, and dynamic gradient-based selection mitigate this by leveraging low-rank projections, ghost gradients, and normalized, periodically refreshed influence measures, while multidimensional schemes like BIDS, DART, IFD, and model-based scoring (AlpaGasus, MoDS) broaden the objective beyond pure loss reduction to incorporate capability balance, difficulty, and alignment quality. This body of work collectively reframes data efficiency as maximizing dynamic marginal utility, not simply removing noise, and it connects directly to compute budgeting by making explicit when the cost of selection outweighs its benefit.

On the memory side, the survey decomposed the training footprint into parameters, optimizer states, and activations, and showed how recent work attacks each term through data-centric coresetting (CoLM, QLESS, Addax), optimizer-centric blockwise updates (HiFT, BAdam), radical zeroth-order and subspace methods (SubZero, LOZO, AdaZeta), and quantization-centric training and adaptation (DQT, Q-LoRA/QA-LoRA, QST, PEQA). While these methods have enabled full-parameter or near–full-parameter fine-tuning of multi-billion-parameter models on commodity hardware, a critical insight of this survey is that almost all of them remain unitary: data-centric strategies primarily relieve activation memory, blockwise optimizers shrink optimizer state, and quantization compresses static model weights, but rarely are these levers combined in a structurally integrated way. The survey thus proposed unified pipelines that simultaneously compress all three memory terms, e.g., blockwise optimization on quantized backbones with zeroth-order regularization for long-context segments, as the most promising route to decouple fine-tuning scale from VRAM limits.

Finally, the compute-governor perspective extended efficiency beyond training mechanics to ask when to ``stop'' or ``reallocate'' compute across data, model size, and inference-time routing. Compute-aware data selection frameworks such as CADS and compute-constrained data selection explicitly incorporate both training and selection overhead into a single budget, establishing payback thresholds where expensive influence- or teacher-based scoring is only beneficial above certain FLOP regimes, and highlighting that “optimal” data subsets are budget-dependent rather than fixed. In parallel, scaling-law work (Kaplan, Chinchilla, data-constrained scaling) provides principled stopping criteria tied to compute-optimal frontiers, clarifying when further epochs or parameter growth yield sharply diminishing returns under a fixed FLOP budget. Budget-aware inference methods, Duo-LLM, Mixture-of-Depths, LayerSkip, MoE routing, and speculative decoding with cascades such as FrugalGPT and distillation-based drafters, extend the same marginal-utility view to decoding, treating per-token depth, expert activation, and verification as decisions governed by latency and cost constraints rather than fixed architectures.

Taken together, the literature surveyed here motivates a unified research agenda for LLM efficiency grounded in three principles: (i) dynamic, compute-aware data valuation that respects both selection and training costs; (ii) holistic memory compression that jointly targets parameters, optimizer states, and activations instead of optimizing a single term in isolation; and (iii) explicit compute governance, in both training and inference, where stopping and routing decisions are framed as marginal-gain-per-FLOP problems rather than ad hoc heuristics. Advancing this agenda will require new algorithmic abstractions that enable true data-memory-compute co-design, for example, extending our compute governor to a joint policy $\pi(S_t, B_t)$ that simultaneously optimizes data subset, memory optimization strategy and model compression, and compute allocation via unified marginal utility $G_t$ across all three pillars, as well as benchmarks and reporting standards that measure not just accuracy but full-stack efficiency across the data–memory–compute triangle. 

If successful, such a perspective could make it possible to train and align frontier-scale models on modest hardware, democratizing access to powerful language models while aligning their development with the environmental and economic constraints of real-world deployment.


In summary, bridging the \textit{Static-to-Dynamic Gap} emerges as a primary frontier for resource-constrained LLM training. Our synthesis suggests that future work must move beyond isolated selection metrics toward \textit{stability-aware dynamic systems}. By addressing the identified technical hurdles, specifically noise-tolerant influence estimation and damped feedback loops, researchers can prevent the ``chatter'' effect in compute-governed pipelines, enabling truly adaptive learning under fixed FLOP budgets. Ultimately, the transition from feasibility-driven optimizations to a unified, budget-aware governance framework will be essential for making LLMs sustainable and accessible in edge-computing environments.

\bibliographystyle{IEEEtran}
\bibliography{Ref}

\end{document}

%% file: finetuningLLM_tikz.tex
\begin{tikzpicture}[
        x=1cm, y=1cm,
        font=\sffamily\scriptsize,
        node distance=0.6cm and 0.4cm,
        base/.style={
            rectangle, 
            rounded corners, 
            draw, 
            align=center, 
            minimum height=1.2cm, 
            minimum width=2.2cm, 
            line width=0.7pt
        },
        subnode/.style={
            rectangle, 
            draw, 
            fill=blue!10!white, 
            align=center, 
            minimum height=1.0cm, 
            minimum width=1.8cm, 
            font=\sffamily\scriptsize, 
            line width=0.5pt
        },
        arrow/.style={-Stealth, thick}
    ]

    \node [base, fill=red!5!white, draw=red!70!black] (llm) {LLM};
    \node [base, fill=blue!5!white, draw=blue!70!black, right=0.8cm of llm] (ft) {Fine-tuning\\the model};
    \node [base, fill=green!5!white, draw=green!70!black, right=0.8cm of ft] (final) {Final Model};

    \node [above=0.4cm of llm] (data1) {Large scale data};
    \node [above=0.4cm of ft] (data2) {Task-specific dataset};

    \node [subnode, below=0.7cm of ft, xshift=-0.9cm] (ift) {Instruction\\Fine-Tuning (IFT)};
    \node [subnode, left=0.2cm of ift] (sft) {Supervised\\Fine-Tuning (SFT)};
    \node [subnode, right=0.2cm of ift] (peft) {PEFT};
    \node [subnode, right=0.2cm of peft] (rlhf) {RLHF};

    \draw [arrow] (data1) -- (llm);
    \draw [arrow] (data2) -- (ft);
    \draw [arrow] (llm) -- (ft);
    \draw [arrow] (ft) -- (final);

    \coordinate (branch) at ($(ft.south) + (0,-0.3)$);
    \draw [thick] (ft.south) -- (branch);
    
    \draw [thick] (sft.north |- branch) -- (rlhf.north |- branch);

    \foreach \x in {sft, ift, peft, rlhf} {
        \draw [arrow] (\x.north |- branch) -- (\x.north);
    }

    \end{tikzpicture}

%% file: quantization.tex
\begin{tikzpicture}[
    node distance=0.5cm,
    font=\sffamily\scriptsize,
    >=Latex,
    frozen/.style={
        rectangle, 
        draw=frozenBorder, 
        fill=frozenBlue, 
        thick, 
        align=center, 
        minimum height=0.8cm, 
        minimum width=2.8cm, 
        rounded corners=2pt,
        inner sep=2pt
    },
    trainable/.style={
        rectangle, 
        draw=trainBorder, 
        fill=trainOrange, 
        thick, 
        align=center, 
        minimum height=0.8cm, 
        minimum width=1.4cm,
        rounded corners=2pt,
        inner sep=2pt
    },
    ghost/.style={
        rectangle,
        draw=gray!40,
        fill=gray!5,
        dashed,
        align=center,
        minimum height=0.8cm,
        minimum width=0.8cm,
        font=\tiny\color{gray}
    },
    flow/.style={
        ->, 
        draw=flowGreen, 
        line width=1pt
    },
    gradient/.style={
        ->, 
        draw=gradRed, 
        dashed, 
        line width=0.8pt
    },
    labeltext/.style={
        font=\bfseries\sffamily\footnotesize, 
        anchor=west
    },
    lossNode/.style={
        align=center, 
        font=\scriptsize, 
        anchor=center
    }
]

\def\vstep{2.5} 

\node[labeltext] at (0,0.4) {A. Direct Quantized Training (DQT)};

\node[frozen, draw=trainBorder, fill=trainOrange] (dqt_weights) at (2.0,-0.3) {$\theta_{int}$ (INT-n)};
\node[ghost, right=2.5cm of dqt_weights] (master) {FP32\\Weights};
\node[lossNode] (lossA) at (5.5,-0.3) {Loss\\$\mathcal{L}$};

\draw[flow] (dqt_weights) -- (lossA) node[midway, above, font=\tiny, color=flowGreen] {Forward};

\node[circle, draw=roundingBorder, fill=roundingGrey, inner sep=0pt, minimum size=0.9cm, align=center] (sr) at (4.5, -1.8) {Stoch.\\Round};

\draw[gradient] (lossA.south) |- (sr.east);
\draw[gradient] (sr.west) -| (dqt_weights.south) node[midway, left, font=\scriptsize, color=gradRed] {Update in-place};

\node[draw=red, thick, circle, minimum size=0.75cm, cross out] at (master.center) {};
\node[font=\tiny, red, align=center, below=0.1cm of master] {No FP32\\Weights saved};

\begin{scope}[yshift=-\vstep cm]
\node[labeltext] at (0,0.4) {B. Q-LoRA / QA-LoRA};

\node[frozen, minimum width=3.5cm] (baseB) at (2.0,-0.4) {Pre-trained Base LLM\\(NF4 Quantized)};

\node[trainable] (lora) at (6.0,-0.4) {LoRA\\Adapter};

\node[lossNode] (lossB) at (8.0,-0.4) {Loss\\$\mathcal{L}$};

\coordinate (inB) at (-0.2,-0.4);

\draw[flow] (inB) -- (baseB.west);
\draw[flow] (baseB.east) -- (lora.west);
\draw[flow] (lora.east) -- (lossB.west);

\draw[gradient] (lossB.south) -- (8.0, -1.4) -| (lora.south);

\node[font=\scriptsize, color=gray!30!black, anchor=north, name=qalora, yshift=-5.5mm, align=center] at (lora.south) {\textbf{QA-LoRA}:~Group-wise Align};
\node[draw=black, circle, inner sep=2pt, fill=white, scale=0.8] (mag) at (qalora.north west) {};
\draw[thick] (mag) -- ++(-0.2,-0.2);

\end{scope}

\begin{scope}[yshift=-2*\vstep cm]
\node[labeltext] at (0,0.4) {C. Quantized Side Tuning (QST)};

\node[frozen, minimum width=3.5cm] (baseC) at (2.0,-0.4) {Frozen LLM\\(INT4)};

\node[trainable, minimum width=1.5cm] (side) at (6.0,-0.4) {Side\\Network};

\node[lossNode] (lossC) at (8.0,-0.4) {Loss\\$\mathcal{L}$};

\coordinate (inC) at (-0.2,-0.4);

\draw[flow] (inC) -- (baseC);
\draw[flow] (baseC) -- (side) node[midway, above, font=\tiny, color=black] {Hidden States};
\draw[flow] (side) -- (lossC);

\draw[gradient] (lossC.south) -- (8.0, -1.4) -| (side.south);

\end{scope}

\begin{scope}[yshift=-2.8*\vstep cm]
\node[labeltext] at (0,0.4) {D. PEQA (Scale Tuning)};

\node[frozen, minimum width=4.0cm] (intMat) at (2.2,-0.4) {Integer Weight Matrix};
\node[trainable, minimum width=1.0cm, anchor=west] (scales) at (intMat.east) {Scales\\$s$};

\node[lossNode] (lossD) at (8.0,-0.4) {Loss\\$\mathcal{L}$};

\draw[decorate, decoration={brace, amplitude=4pt, mirror}, thick, gray, align=center] (intMat.south west) -- (scales.south east) node[midway, below=3pt, font=\scriptsize, color=black] {Parameter decomposition\\of original weight $\theta$};

\coordinate (inD) at (-0.2,-0.4);
\draw[flow] (inD) -- (intMat);
\draw[flow] (scales.east) -- (lossD);

\draw[gradient] (lossD.south) -- (8.0, -1.5) -| (scales.south);

\end{scope}

\node[draw=gray!80, fill=white, rounded corners, anchor=north west, align=left, font=\footnotesize, inner sep=4pt] at (0, -2.7*\vstep - 2.0) {
    \textbf{Legend:} \quad 
    \tikz\node[draw=frozenBorder, fill=frozenBlue, inner sep=2pt, minimum size=8pt, rounded corners=1pt] {}; Frozen \quad
    \tikz\node[draw=trainBorder, fill=trainOrange, inner sep=2pt, minimum size=8pt, rounded corners=1pt] {}; Trainable \quad
    \tikz\draw[flowGreen, ->, thick] (0,0) -- (0.5,0); Fwd \quad
    \tikz\draw[gradRed, ->, dashed, thick] (0,0) -- (0.5,0); Grad
};

\end{tikzpicture}

%% file: governor.tex
\definecolor{govBlue}{RGB}{0, 82, 164}
\definecolor{constrGray}{RGB}{110, 110, 110}
\definecolor{stageGreen}{RGB}{92, 170, 92}
\definecolor{stageOrange}{RGB}{220, 130, 50}
\definecolor{stagePurple}{RGB}{143, 92, 194}
\definecolor{actionLightBlue}{RGB}{225, 240, 255}
\definecolor{lifecycleGroupColor}{RGB}{230, 230, 240}

\begin{tikzpicture}[
    font=\sffamily,
    node distance=0.6cm and 0.8cm,
    basebox/.style={
        rectangle, rounded corners, align=center, thick, minimum width=2.6cm, inner sep=6pt, font=\small\sffamily
    },
    constraint/.style={
        basebox, draw=constrGray, fill=gray!5, minimum height=1.6cm, minimum width=3.2cm
    },
    governor/.style={
        rectangle, rounded corners=15pt, draw=govBlue, fill=govBlue!5, very thick, 
        minimum width=3.5cm, minimum height=5.5cm, align=center 
    },
    actionbtn/.style={
        rectangle, rounded corners, draw=govBlue, fill=actionLightBlue, 
        minimum width=2.8cm, minimum height=0.5cm, align=center, font=\scriptsize\sffamily
    },
    stagePT/.style={basebox, draw=stageGreen!80!black, fill=stageGreen!20, minimum height=2.2cm},
    stageFT/.style={basebox, draw=stageOrange!80!black, fill=stageOrange!20, minimum height=2.2cm},
    stageINF/.style={basebox, draw=stagePurple!80!black, fill=stagePurple!20, minimum height=2.2cm, minimum width=5.6cm},
    inputArrow/.style={->, >=Latex, thick, constrGray},
    controlArrow/.style={->, >=Latex, line width=2pt, govBlue},
    feedbackArrow/.style={->, >=Latex, line width=2pt, constrGray, dashed, rounded corners=10pt},
    labelText/.style={font=\bfseries\small\sffamily, fill=white, inner sep=3pt, align=center, rounded corners},
    lifecycleBlock/.style={
        rectangle, draw=stagePurple!50!black, fill=lifecycleGroupColor, rounded corners, thick, inner sep=10pt, dashed
    }
]

    \node[governor] (gov) at (0, 0) {};

    \node[font=\sffamily\bfseries\normalsize, anchor=north] (govTitle) at ($(gov.north)-(0,0.2)$) {GOVERNOR};
    \node[font=\footnotesize, anchor=north] (govSub) at ($(govTitle.south)-(0,0.05)$) {Policy $\pi(S_t, B_t) \to a_t$};

    \node[actionbtn] (act_cont) at ($(govSub.south)-(0,0.7)$) {{\bfseries CONTINUE}\\[0.1em]{\scriptsize Maintain}};
    \node[actionbtn] (act_reall) [below=0.4cm of act_cont] {{\bfseries REALLOCATE}\\[0.1em]{\scriptsize Switch Strategy}};
    \node[actionbtn] (act_stop) [below=0.4cm of act_reall] {{\bfseries STOP}\\[0.1em]{\scriptsize Exit/Offload}};

    \node[draw=govBlue, fill=white, rounded corners, font=\scriptsize\sffamily, anchor=north, inner sep=4pt, yshift=8pt]
        (metric) at (gov.south) {$G_t = \frac{-\Delta\mathcal{L}}{\Delta\text{FLOPs}}$};

    \node[constraint] (budget) [left=1.3cm of gov, yshift=1.5cm] {{\bfseries\footnotesize BUDGET $B_t$}\\[0.2em]{\scriptsize FLOPs, Time}};
    \node[constraint] (context) [left=of gov, yshift=-1cm] {\parbox{4cm}{\centering
        {\bfseries\footnotesize RESOURCES INTERDEPENDENCY}\\[0.2em]
        {\scriptsize Coupled Data-Memory-Compute Trade-offs \\ Navigating the Pareto front where resource gains (e.g., Quantization or Pruning) incur overhead "taxes" in other dimensions. }
    }};

    
    \node[stagePT] (pt) [right=1cm of gov.north east, anchor=north west, yshift=-0.5cm] {\parbox{3cm}{\centering
    {\bfseries\footnotesize PRE-TRAINING}\\[0.2em]{\scriptsize Parameter-
Heavy~\cite{kaplan2020scaling}, Data-Rich~\cite{hoffmann2022training}, and Data-Constrained Regimes~\cite{muennighoff2023scaling}
    }}};
    
    \node[stageFT] (ft) [right=0.4cm of pt] {\parbox{3cm}{\centering
    {{\bfseries\footnotesize FINE-TUNING}\\[0.2em]{\scriptsize         CADS~\cite{sachdeva2024traindataefficientllms},\\ Payoff-Optimal Data Selection~\cite{yincompute}}}}};
    
    \node[stageINF] (inf) at ($(pt.south west)!0.5!(ft.south east) + (0, -0.6)$) [anchor=north] {\parbox{6cm}{\centering
    {\bfseries\footnotesize INFERENCE}\\[0.2em]
    {\scriptsize Per-Layer Routing:              Duo-LLM~\cite{alizadeh2024duo}, \\Per-Depth Skipping:             MoD~\cite{raposo2024mixtureofdepthsdynamicallyallocatingcompute}, LayerSkip~\cite{Elhoushi_2024}\\
 Architectural MoE:              Mixtral~\cite{jiang2024mixtralexperts}, DeepSeekMoE~\cite{dai2024deepseekmoeultimateexpertspecialization}\\
Decoding Cascades:              FrugalGPT~\cite{chen2023frugalgpt}, DistillSpec~\cite{zhou2023distillspec}}
    }};

    \begin{pgfonlayer}{background}
        \node[fit=(budget)(context), draw=constrGray!50, dashed, rounded corners, fill=gray!2, inner sep=8pt, label={[constrGray!80, yshift=2pt, font=\bfseries\footnotesize]above:Constraints}] (inputGroup) {};
        
        \node[lifecycleBlock, fit=(pt)(ft)(inf)] (lifecycleGroup) {};
    \end{pgfonlayer}

    \draw[inputArrow] (budget.east) -- (gov.west |- budget.east);
    \draw[inputArrow] (context.east) -- (gov.west |- context.east);
    
    \draw[controlArrow] (gov.east) -- node[labelText, above, yshift=3pt] {$a_t$} (gov.east -| lifecycleGroup.west);

    \draw[feedbackArrow] (lifecycleGroup.north) -- ++(0,0.8) -| (gov.north)
        node[pos=0.25, labelText] {$S_t$};

\end{tikzpicture}